\documentclass[journal]{IEEEtran}
\ifCLASSOPTIONcompsoc
  \usepackage[nocompress]{cite}
\else
  \usepackage{cite}
\fi
\usepackage{amsmath}
\usepackage{amssymb}
\usepackage{url}
\usepackage{subfig}
\usepackage{arydshln}
\usepackage{amsfonts}
\usepackage{enumerate}
\usepackage{epsfig}
\usepackage{graphicx}
\usepackage{balance}
\usepackage{hyperref}

\usepackage[noend]{algpseudocode}
\usepackage{algorithmicx,algorithm}

\def\ie{{\em i.e.}}

\def\etal{{\em et al.}}

\newcommand{\secref}[1]{Section \ref{#1}}

\graphicspath{{./figure/}}

\usepackage{ragged2e}
\usepackage{booktabs}
\usepackage{color}
\usepackage{multirow}
\usepackage{bm}
\usepackage{bbm}

\begin{document}

\title{Assessment of Deep Learning-based Heart Rate Estimation using Remote Photoplethysmography under Different Illuminations}
 
\author{Ze~Yang,~Haofei~Wang,~\IEEEmembership{Member,~IEEE}, and~Feng~Lu,~\IEEEmembership{Member,~IEEE}
\IEEEcompsocitemizethanks{\IEEEcompsocthanksitem Ze Yang and Feng Lu are with the State Key Laboratory of Virtual Reality Technology and Systems, School of Computer Science and Engineering, Beihang University, Beijing, 100191, China. (email:{yangze, lufeng}@buaa.edu.cn).
\IEEEcompsocthanksitem Haofei Wang is with Peng Cheng Laboratory, Shenzhen, 518055, China.
(email: wanghf@pcl.ac.cn)
\IEEEcompsocthanksitem Feng Lu is the corresponding author.}
}% <-this % stops a space

\maketitle

\begin{abstract}
  Remote photoplethysmography (rPPG) monitors heart rate without requiring physical contact, which allows for a wide variety of applications. Deep learning-based rPPG have demonstrated superior performance over the traditional approaches in controlled context. However, the lighting situation in indoor space is typically complex, with uneven light distribution and frequent variations in illumination. It lacks a fair comparison of different methods under different illuminations using the same dataset. In this paper, we present a public dataset, namely the BH-rPPG dataset, which contains data from thirty five subjects under three illuminations: low, medium, and high illumination. We also provide the ground truth heart rate measured by an oximeter. We evaluate the performance of three deep learning-based methods (Deepphys, rPPGNet, and Physnet) to that of four traditional methods (CHROM, GREEN, ICA, and POS) using two public datasets: the UBFC-rPPG dataset and the BH-rPPG dataset. The experimental results demonstrate that traditional methods are generally more resistant to fluctuating illuminations. We found that the Physnet achieves lowest mean absolute error (MAE) among deep learning-based method under medium illumination, whereas the CHROM achieves 1.04 beats per minute (BPM), outperforming the Physnet by 80$\%$. Additionally, we investigate potential methods for improving performance of deep learning-based methods. We find that brightness augmentation make model more robust to variation illumination. These findings suggest that while developing deep learning-based heart rate estimation algorithms, illumination variation should be taken into account. This work serves as a benchmark for rPPG performance evaluation and it opens a pathway for future investigation into deep learning-based rPPG under illumination variations.
\end{abstract}


% \fi
\IEEEpeerreviewmaketitle

%asdfadsfasdasdf
\section{Introduction}\label{sec:intro}

\IEEEPARstart{H}{eart} rate (HR) is an important physiological indicator for both physical and mental health. HR monitoring has been used in many applications, such as state monitoring~\cite{statemonitoring}, driver fatigue detection \cite{driver}, face anti-spoofing \cite{FaceLivenessDetection}, etc. Traditional HR monitoring methods rely on electrocardiograph (ECG) and contact photoplethysmography (PPG) sensors. However, wearing such contact devices is uncomfortable and often interferes with daily activities. With the development of computer vision algorithm, remote HR measurement based on remote photoplethysmograply (rPPG) has been proposed~\cite{6523142,chenDeepphysVideobasedPhysiological2018,iphys,spetlikVisualHeartRate2018}. While rPPG offers the potential for contactless and continuous measurement of HR using low-cost web cameras, the system performance is still limited by many factors, such as lighting variations and head movements\cite{liRemoteHeartRate2014}.

Lighting conditions is critical for rPPG since the quality of rPPG signal is determined by the light ingested into the skin. However, most existing studies only explored the laboratory condition with good lighting condition.~\cite{chenDeepphysVideobasedPhysiological2018,yuRemoteHeartRate2019,6523142,verkruysseRemotePlethysmographicImaging2008,NoncontactAutomatedCardiac,wangAlgorithmicPrinciplesRemote2016}. 
Insufficient illumination may lead to low amplitude of rPPG signal, due to the fact that the energy of light is too vulnerable to penetrate into skin surface. Moreover, most traditional methods are required to find the specific skin area \cite{wangAlgorithmicPrinciplesRemote2016, 6523142} and yet the low contrast on an image makes it difficult to obtain correct region of interest (ROI). On the contrary, high intensity of light lead to image clipping on the skin surface~\cite{wangRobustAutomaticRemote2017,papageorgiou2014adaptive}. Besides, the light distribution also has significant impact on rPPG. The conventional methods usually select the whole face as ROI and assumes the same contribution of rPPG signal at different parts of the face. This assumption may not hold in the real-world applications, especially in indoor space, the light distribution and intensity are different due to the relative position between the subject and the light source.

The traditional approaches use different methods to extract rPPG signal, which can be mainly categorized into two types: 1) skin reflection model-based approach\cite{6523142,wangAlgorithmicPrinciplesRemote2016} and 2) blind source separation-based approach \cite{NoncontactAutomatedCardiac,PCA}. Unfortunately, these models seldom take the lighting conditions into consideration. Po \etal{} proposed an adaptive ROI approach-based on the quality of rPPG signal acquired from sub-region of face to tackle the uneven light distribution challenge. However, the system performance under different lighting intensities has not yet been evaluated.

Deep learning-based approaches, such as Convolutional Neural Networks (CNN), have been used to estimate HR using rPPG signal. Špetlik\etal{} \cite{spetlikVisualHeartRate2018} proposed to use 2D-CNN as backbone to directly estimate single value of HR at the early stage of rPPG, but it neglects the temporal information between frames. Physnet\cite{yuRemotePhotoplethysmographSignal2019} and rPPGNet \cite{yuRemoteHeartRate2019} can detect atrial fibrillation (AF) by generating more precise rPPG signals. The Physnet employs deep spatio-temporal networks with 3D-CNN as backbone and builds an end-to-end model. The rPPGNet treats the HR estimation as multi-task learning task (HR estimation task and skin segmentation task), which also uses the 3D-CNN as backbone with skin segmentation branch. Although these works achieve the superior performance, the quality of rPPG signal generated under different illumination is still unknown. The Deepphys\cite{chenDeepphysVideobasedPhysiological2018} combines the theory of skin reflection model and attention mechanism, which adopt 2D-CNN as backbone. It outperforms the traditional methods by using attention mechanism, which takes into account the rPPG intensity distribution in different parts of the face. It is well-known that the performance of deep learning model is sensitive to illumination. While the filters in CNN model learn the specific pattern capturing different levels of visual information in most of computer vision tasks, the lighting affects the quality of the rPPG signal itself in HR estimation task. Wang \etal{}\cite{zhan2020analysis} conducted a series of experiments to show that CNN uses color variation information in blood absorption to estimate HR. However, they did not validate the performance of deep learning models under different illuminations.

For data-driven approaches, the quality of training data determines the system performance. Most of the previous studies evaluated the performance on different datasets, which makes it unfair to compare the system performance. For example, the Physnet\cite{yuRemotePhotoplethysmographSignal2019} is trained on OBF\cite{liOBFDatabaseLarge2018} dataset, the Deepphys is trained on RGB Video I\cite{chenDeepphysVideobasedPhysiological2018} dataset, both of the datasets are not publicly accessible.

\begin{figure}[!t]
	\centering
	\includegraphics[width= 0.5\textwidth]{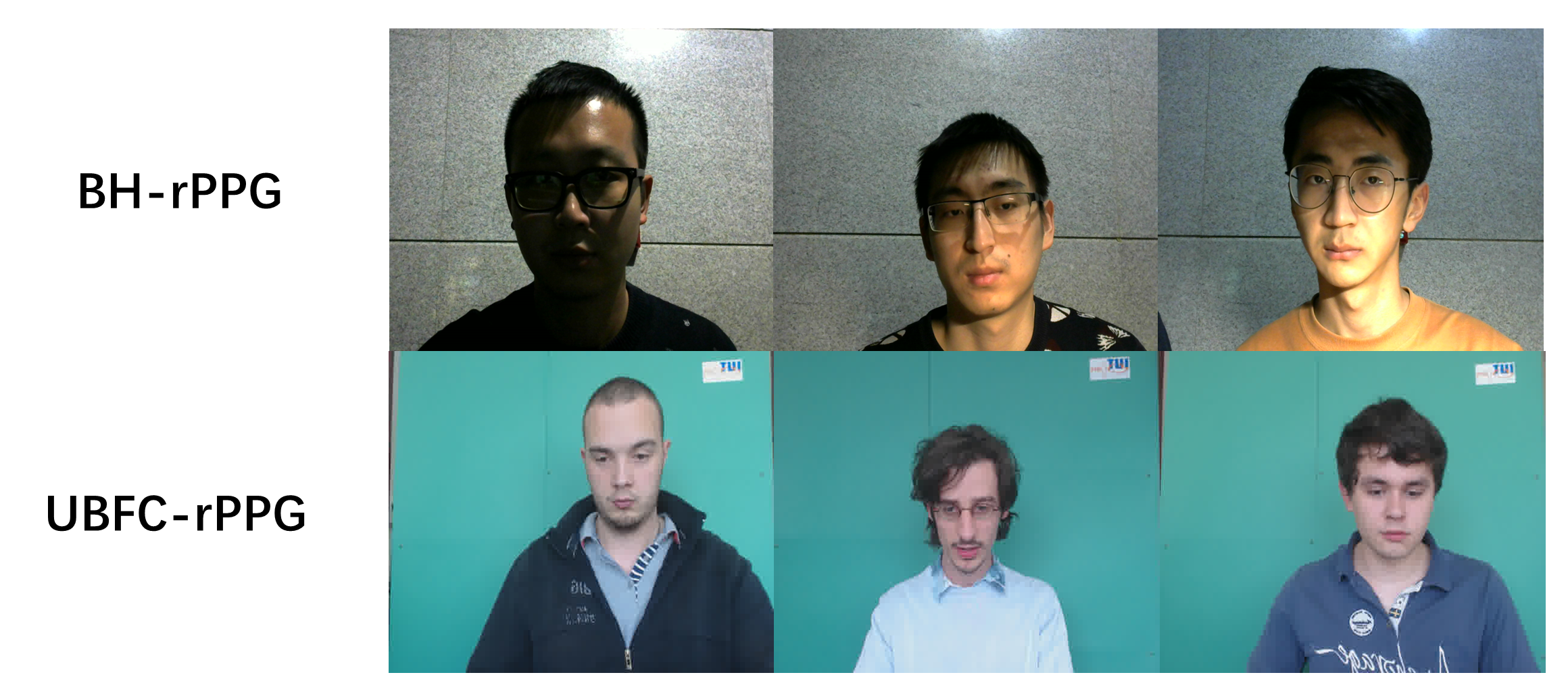}
	\caption{Uncontrolled light sources and controlled light sources. The first row is collected in natural lighting condition. The lighting intensity on face is uneven and the three columns from left to right correspond to pictures taken at different light intensities. The second row represents video recorded in controlled lighting environment.}
	\label{dataoflight}
\end{figure}

To evaluate the robustness of different methods in real-world applications, here we presented a public dataset, i.e., BH-rPPG dataset (BH stands for BeiHang University), which consists of three lighting intensities with uneven light distribution on the face (see first row in Fig.\ref{dataoflight}). 

In summary, the primary contributions of this paper are three-fold:
\begin{enumerate}

\item We present the BH-rPPG, a public dataset for rPPG-based heart rate estimation. The BH-rPPG consists of twelve subjects' data under three different illuminations. The link can be found at \href{https://github.com/yangze68/BH-rPPG}{https://github.com/yangze68/BH-rPPG}.
\item We systematically evaluated the robustness to illumination variation of typical methods for rPPG-based heart rate estimation, including four traditional methods \cite{6523142,verkruysseRemotePlethysmographicImaging2008,NoncontactAutomatedCardiac,wangAlgorithmicPrinciplesRemote2016}) and three deep learning-based methods ~\cite{chenDeepphysVideobasedPhysiological2018,yuRemoteHeartRate2019,yuRemotePhotoplethysmographSignal2019}).
\item Our experimental results suggest that although the deep learning-based methods achieve superior performance under normal illumination, they are less resistant to illumination variations compared with traditional methods. We also explore potential methods for performance improvement using the deep learning-based model. Results show that brightness augmentation is efficient in improving performance in different lighting conditions. These findings draw attention to designing more robust deep learning-based methods for remote heart rate estimation.
\end{enumerate}

The remainder of this paper is organized as follows.~\secref{sec:relatedwork} summarizes the related works of the heart rate estimation methods using rPPG and the lighting conditions in different applications.~\secref{sec:datasets} describes the datasets including the UBFC-rPPG and the proposed BH-rPPG.~\secref{sec:experiment setup} describes the experimental setup including methods, experimental protocols, performance evaluation metrics and potential method for performance improvement.~\secref{sec:experiment results} presents the experimental results.~\secref{sec:discussion} gives a discussion of the findings. Finally,~\secref{sec:conclusion} concludes this paper and outlines the future work.

\section{Related Work}\label{sec:relatedwork}
% We begin with a general introduction  of the diverse methods of rPPG-based applications, followed by previous methods for rPPG.
We first review the existing methods of heart rate estimation using rPPG, including both traditional approaches and deep learning-based approaches. Then we summarize the different lighting conditions in various rPPG applications.

\subsection{Heart Rate Estimation via rPPG}\label{sec:method}
Heart rate can be remotely monitored through two channels: ballistocardiographic (BCG)\cite{balakrishnanDetectingPulseHead2013,ImprovedPulseDetection} and remote photoplethysmography (rPPG) \cite{chenDeepphysVideobasedPhysiological2018,yuRemoteHeartRate2019,6523142,verkruysseRemotePlethysmographicImaging2008,NoncontactAutomatedCardiac,wangAlgorithmicPrinciplesRemote2016}. 
The BCG-based methods use a camera to capture subtle movements induced by the periodic blood ejected into the vessels with each heartbeat. The BCG-based non-contact pulse measurements is achieved by blind source separation of the head movements in video. However, BCG-based methods are usually limited by user's head movement since the faint movements trace induced by cardiac activity is hard to capture during large scale head movements. On the contrary, rPPG-based methods register the pulse induced by subtle color variations of human skin\cite{takanoHeartRateMeasurement2007,verkruysseRemotePlethysmographicImaging2008}. This measurement is based on the fact that the pulsatile blood propagating in the human cardiovascular system changes the blood volume in skin tissue. The oxygenated blood circulation leads to fluctuations in the amount of hemoglobin molecules and proteins thereby causing variations in the optical absorption and scattering across the light spectrum\cite{takanoHeartRateMeasurement2007}.

The rPPG-based methods can be categorized into two types, 1) the traditional methods that rely on optical models, e.g., the Lambert-Beer law and the Shafer's dichromatic reﬂection model; and 2) the deep learning-based methods that rely on the appearance of the face.

The optical models that used in the traditional methods are grounded by the optical properties of the skin under ambient illumination. Different color channels contain different quality of rPPG signal. The green channel has been used in early rPPG research, since it generates strongest rPPG signal \cite{verkruysseRemotePlethysmographicImaging2008}. Previous studies have shown that the cardiac activity causes variation in the optical absorption across the light spectrum\cite{allenPhotoplethysmographyItsApplication2007}, using this characteristic, CHROM\cite{6523142} and POS\cite{wangAlgorithmicPrinciplesRemote2016} project RGB on different plane by re-weighting and linear combination of color channel. Blind signal separation has also been proposed, which considers the temporal trace of PPG that can be retrieved from independent or uncorrelated signal sources under certain assumptions. Independent component analysis has been used in multi-signal sources obtained by different approaches such as the same region of color channels \cite{NoncontactAutomatedCardiac} and patch level of regions of interest (ROI) \cite{lamRobustHeartRate2015}.

Many deep learning-based heart rate estimation methods have been proposed recently. Chen and McDuff \cite{chenDeepphysVideobasedPhysiological2018} present Deepphys, which employs two parallel branches of CNN to extract rPPG feature: the motion branch and the appearance branch. The motion branch is fed with normalized frame-differences to cancel motion effect on rPPG signal, while the appearance branch uses attention mechanism that enables the network to focus on the area of skin. Other researchers investigated different network architectures for better estimation. Yu et al. \cite{yuRemotePhotoplethysmographSignal2019} developed an end-to-end network to estimate heart rate using compressed videos. They used a three dimensional CNN to capture temporal information and an extra skin segmentation branch to regress PPG signal. Niu et al.\cite{niuRhythmnetEndtoendHeart2019} proposed to directly estimate heart rate from a spatio-temporal network. % Since the human face is not covered by clothing and is exposed to the camera for a long period of time, it is more appropriate to extract the rPPG for the face.
Špetlik et al.\cite{spetlikVisualHeartRate2018} introduce two-step network for feature extraction and heart rate estimation. Qiu et al.\cite{qiuEVMCNNRealtimeContactless2018} integrated the signal magnified technique named Eulerian video magnication\cite{wuEulerianVideoMagnification2012} with convolutional neural network to estimate heart rate.
Lee et al.\cite{MetarPPGRemoteHeart} proposed a transductive meta-learner to adapt model to different domains. In addition to meta-learner, Niu et al. \cite{VideoBasedRemotePhysiological} introduced a cross-verified scheme to purify the feature constructed with spatio-temporal map. Although deep learning methods yield promising results, their performance under different illumination remains to be explored.

\subsection{Lighting conditions in rPPG applications}
% The application field for rPPG can be divided into two categories.
% State monitoring is the most intuitive rPPG applications, including clinical health monitoring and home health monitoring. 
% The clinical health monitoring is that it can replace obtrusive sensors such as Electrocardiography (ECG) and Photoplethysmography (PPG), in which the electrode attached to the patient skin for cardiac activity and pulse-rate monitoring respectively. This is especially beneficial for patients with fragile skin (e.g., burn patient) in a Intensive Care Unit (ICU).
% Moreover, with the number of coronavirus disease cases in 2019 (COVID-19) is increasingly growing, non-contact healthcare techniques can contribute to diagnosis\cite{smithTelehealthGlobalEmergencies2020}.
The rPPG has been used in many applications, such as state monitoring at home or driver fatigue detection on the car, where the lighting conditions can be very different.

In indoor environment, depending on the relative position between the person and light source, it may suffer from insufficient and uneven lighting conditions. In the application of state monitoring, the algorithm should adapt to the light variation. For example, Sun \etal{}\cite{sun2021camera} continuously monitored discomforts of infants over a long period, which requires the algorithm to work in complex light conditions. To estimate heart rate in extremely low light condition, Lin \etal{}\cite{linusing} proposed to use infrared spectrum to extract features. In addition, due to the COVID-19, the demand for non-contact healthcare techniques is dramatically increasingly \cite{smithTelehealthGlobalEmergencies2020}. However, the investigation of algorithm performance under complex lighting conditions is relatively rare.

In outdoor environment, the heart rate estimation becomes more challenging since the illumination changes dramatically \cite{huang2020heart}. In the driver fatigue detection task, the illumination is quite distinct from the laboratory. In other applications such as face anti-spoofing and online payment system, rPPG technology could be used to perform liveness detection, which prevents using a fake face to circumvent the system and to gain unauthorized access\cite{FaceLivenessDetection}. The Deepfake video can also be distinguishable by rPPG\cite{fernandesPredictingHeartRate2019}, whereas the lighting conditions is far more complicated due to wide range of usage.

In summary, although rPPG has been deployed in many applications, the non-ideal lighting conditions degrade the system performance. Thus, it is necessary to conduct a systematic comparison between different approaches and to evaluate the robustness of these approaches under different lighting conditions.
\section{Datasets}\label{sec:datasets}
In this section, we first briefly introduce the public dataset used in the experiment. Then, we present the details of BH-rPPG dataset under three different lighting conditions: low/medium/high illumination.

\subsection{Public dataset}
Most public datasets are collected under controlled environment, such as UBFC-rPPG~\cite{ubfc-rppg}, VIPL~\cite{VIPL}, PURE~\cite{PURE} and MAHNOB~\cite{MAHNOB}. To the best of our knowledge, no public dataset examines the effect of illumination intensities. Although COHFACE \cite{cohface} dataset was collected under controlled lighting and natural lighting, the lighting intensity remains the same.
Here we choose the UBFC-rPPG \cite{ubfc-rppg} dataset as the training set for deep learning method.

The University Bourgogne Franche-Comté Remote PhotoPlethysmoGraphy dataset (\ie, the UBFC-rPPG dataset) consists of two scenarios, here we only use the part that subjects play a time-sensitive mathematical game. This is because it is a real life setting which includes natural head movements. Subjects' heart rate changes over time as induced by mathematical games. The dataset includes 42 one-minute videos from different subjects. The video is recorded using a low-cost webcam (Logitech C920 HD Pro) at 30fps with a resolution of 640x480 in uncompressed 8-bit RGB format. A CMS50E transmissive pulse oximeter was used to obtain the ground truth PPG data comprising of the PPG waveform as well as the PPG heart rates.

% For evaluation of deep learning method, the training set contains 37 subjects and the testing set contains the rest 5 subjects.

\subsection{BH-rPPG dataset}
\subsubsection{Apparatus setup}

\begin{figure}[!t]
	\centering
	\includegraphics[width=0.4\textwidth]{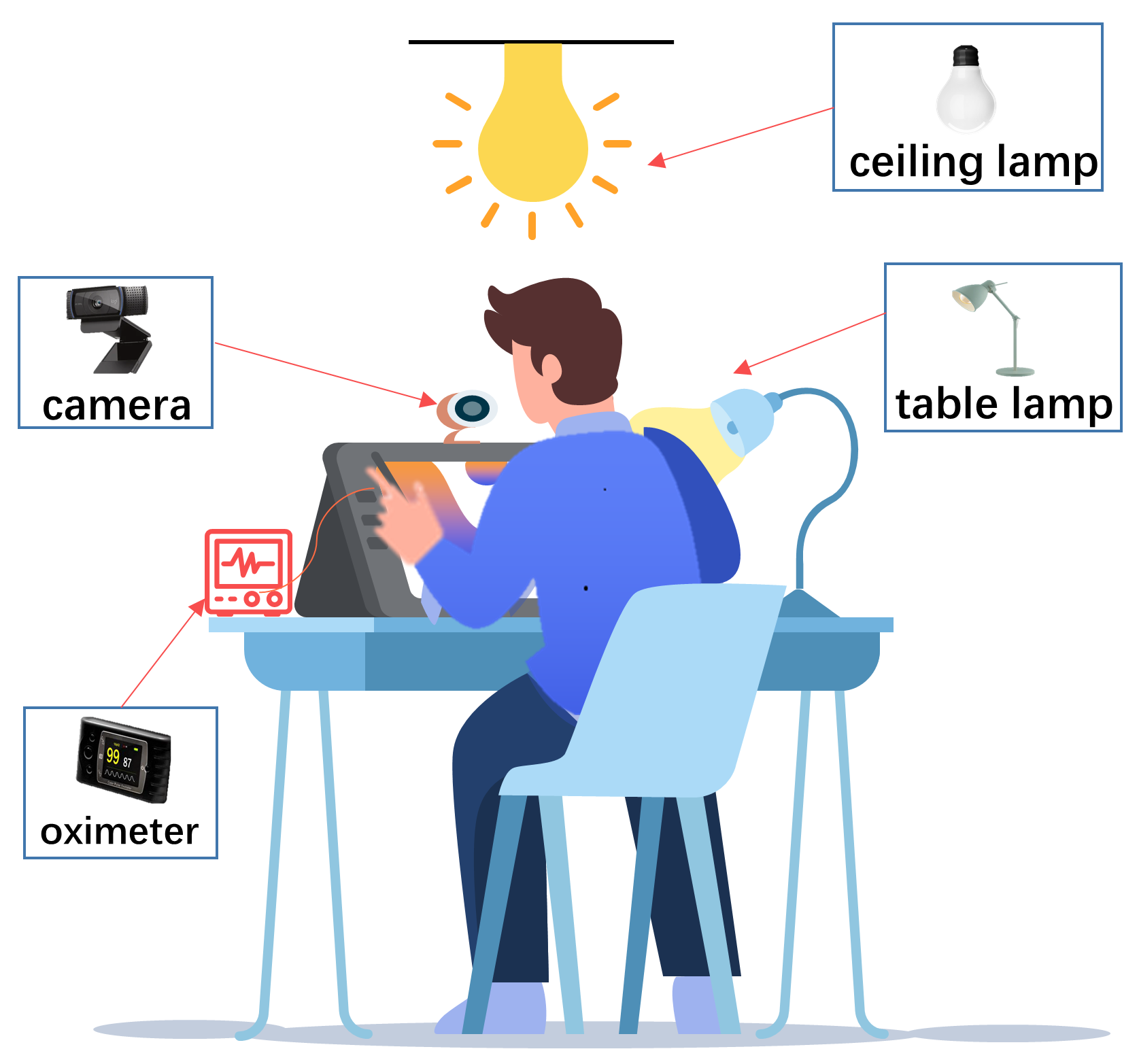}
	\caption{Experimental setup.}
	\label{experimentfig}
\end{figure}

Fig.\ref{experimentfig} presents the experimental setup. There are two light sources (a ceiling lamp and a table lamp) that create different lighting conditions. An oximeter (CONTEC CMS50E) was used to obtain the ground truth PPG data. A webcam (Logitech HD pro webcam C310 color camera) recorded the video data synchronized with the oximeter. The resolution for video is $640\times480$. The web camera actual frame rate is 30 fps, but under low lighting intensity, the actual frame rate is about 20 fps. The subject sits 1 meter away from the camera. Since our study focuses on illumination variations instead of head movements, subjects are asked to keep their head stationary during the data collection. The reason that we used two lamps in the experiment is that this is more similar to the settings in daily living. We collected data under three lighting conditions, as shown in Table \ref{datasettable}. With the ceiling lamp always on in three conditions, we change the mode of table lamp to modulate the illumination. Fig. \ref{lighting_intensity} shows some sample images under different illuminations.

\begin{table}[h]
  \caption{Three lighting conditions in our dataset.}
  \label{datasettable}
  \begin{center}
  \begin{tabular}{ccccccc}
  \hline
  Lighting Condition& Ceiling Lamp& Table Lamp\\
  \hline
  Low intensity level& $\surd$& off\\
  Medium intensity level& $\surd$& normal\\
  High intensity level& $\surd$& high\\
  \hline
  \end{tabular}
  \end{center}
  \end{table}
  
\subsubsection{Data collection procedure}

\begin{figure}[!t]
	\centering
	\includegraphics[width= 0.4\textwidth]{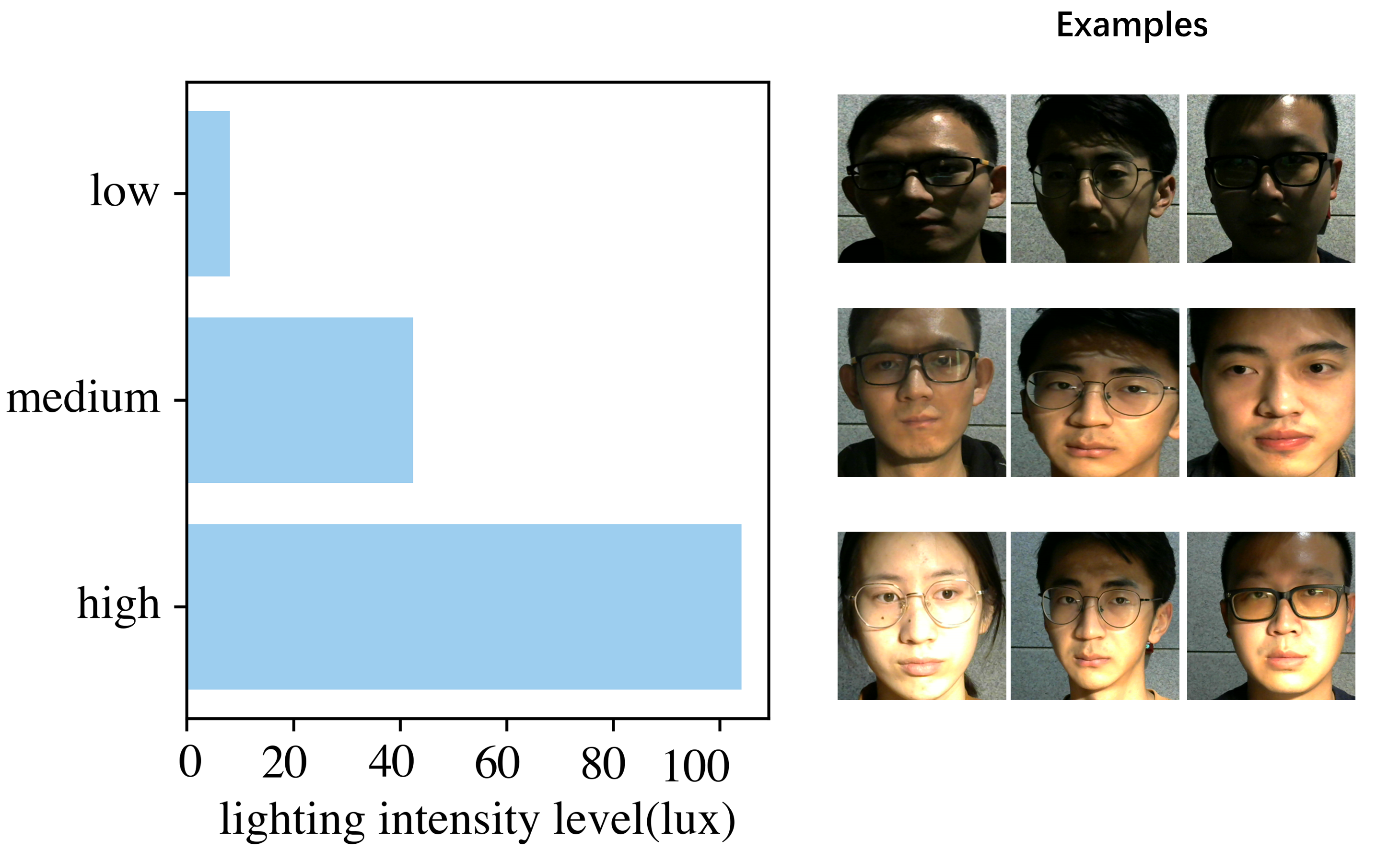}
	\caption{Average illumination and sample images.}
	\label{lighting_intensity}
\end{figure}

We recruited 35 healthy subjects (16 males and 19 females, with a gender ratio of 0.84) on campus, with a mean age of 24, SD of 2.31. For each subject, we took three 30-second videos under three lighting conditions. The left part of Fig.\ref{lighting_intensity} shows the average lighting intensity under three conditions. The illuminations of low, medium and high level are 8.0, 42.4, and 104.0 lux, respectively.
\subsubsection{Dataset statistics}
\begin{figure}[!t]
	\centering
	\includegraphics[width= 0.4\textwidth]{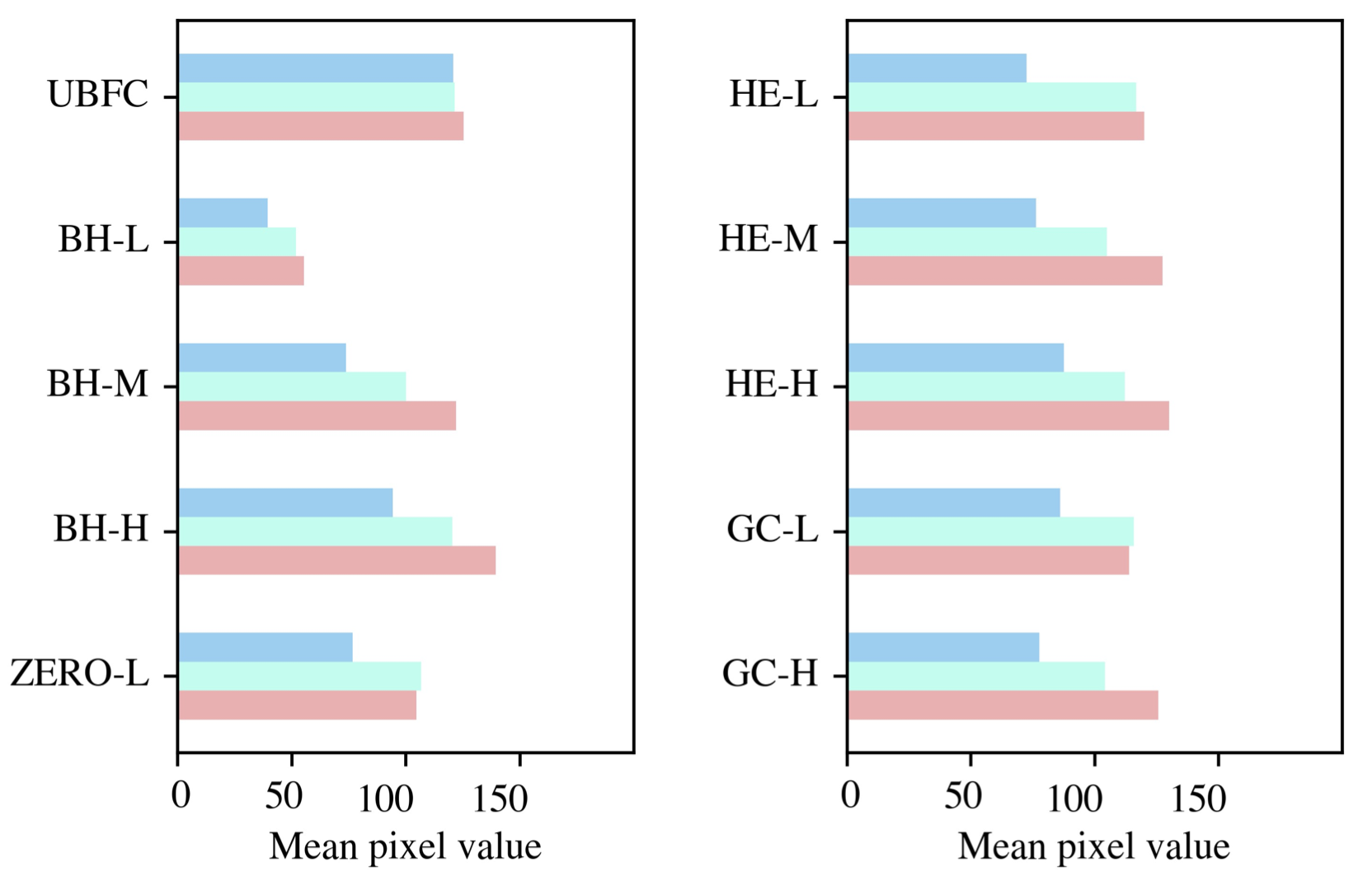}
	\caption{RGB mean values of different dataset. BH represents BH-rPPG. L, M and H represents the lighting conditions. ZERO, HE and GC represents the enhanced output of BH-rPPG, which is enhanced by ZERO-DCE, HE and GC respectively. The red, green and blue bar represents the RGB mean values of each dataset.}
	\label{dataset_rgb}
\end{figure}

The BH-rPPG dataset consists of 105 videos from 35 participants(refer to Table \ref{datasetstatistic}.) To quantitatively demonstrate the variations in lighting in the BH-rPPG dataset, we also calculated the
bar charts of the mean value of videos in different lighting conditions. Due to the fact that we only care about the light conditions on face area, we compute the mean pixel within the region of the bounding box of face (as shown in Fig. \ref{dataset_rgb}).
\begin{table}[h]
\centering
\caption{A summary of the UBFC-rPPG and our BH-rPPG datasets for statistics difference}
\label{datasetstatistic}
\begin{tabular}{lcc} 
\hline
              & UBFC                               & BH-rPPG          \\ 
\hline
Subjects      & 42                                 & 35               \\
Illumination  & Lab environment                    & Low/Medium/high  \\
Head movement & \multicolumn{1}{l}{Large movement} & stable           \\
Videos        & 42                                 & 105              \\
Frame rate    & 30fps                              & about 20fps      \\
\hline
\end{tabular}
\end{table}

\section{Experimental setup}\label{sec:experiment setup}
In this section, we introduce the methods compared in this paper. Next, we describe our experimental protocol and the performance evaluation metrics. Finally, to gain further understanding of the illumination effect on HR estimation accuracy, we investigate several potential methods, such as illumination compensation techniques and training strategies, to promote the performance of HR estimation with current deep learning-based models.
\subsection{Methods}
We evaluate both traditional methods and deep learning-based methods. To make a fair comparison and eliminate the difference during preprocessing, we used Viola-Jones face detector to extract face area for reducing noise from the background. We employed Kanade-Lucas-Tomasi (KLT)~\cite{KLT} algorithm to track the location of the face region to avoid head rigid movements. The processed video frames are used as input for different algorithms.

\subsubsection{Traditional methods}
\begin{figure}[!t]
	\centering
	\includegraphics[width= 0.5\textwidth]{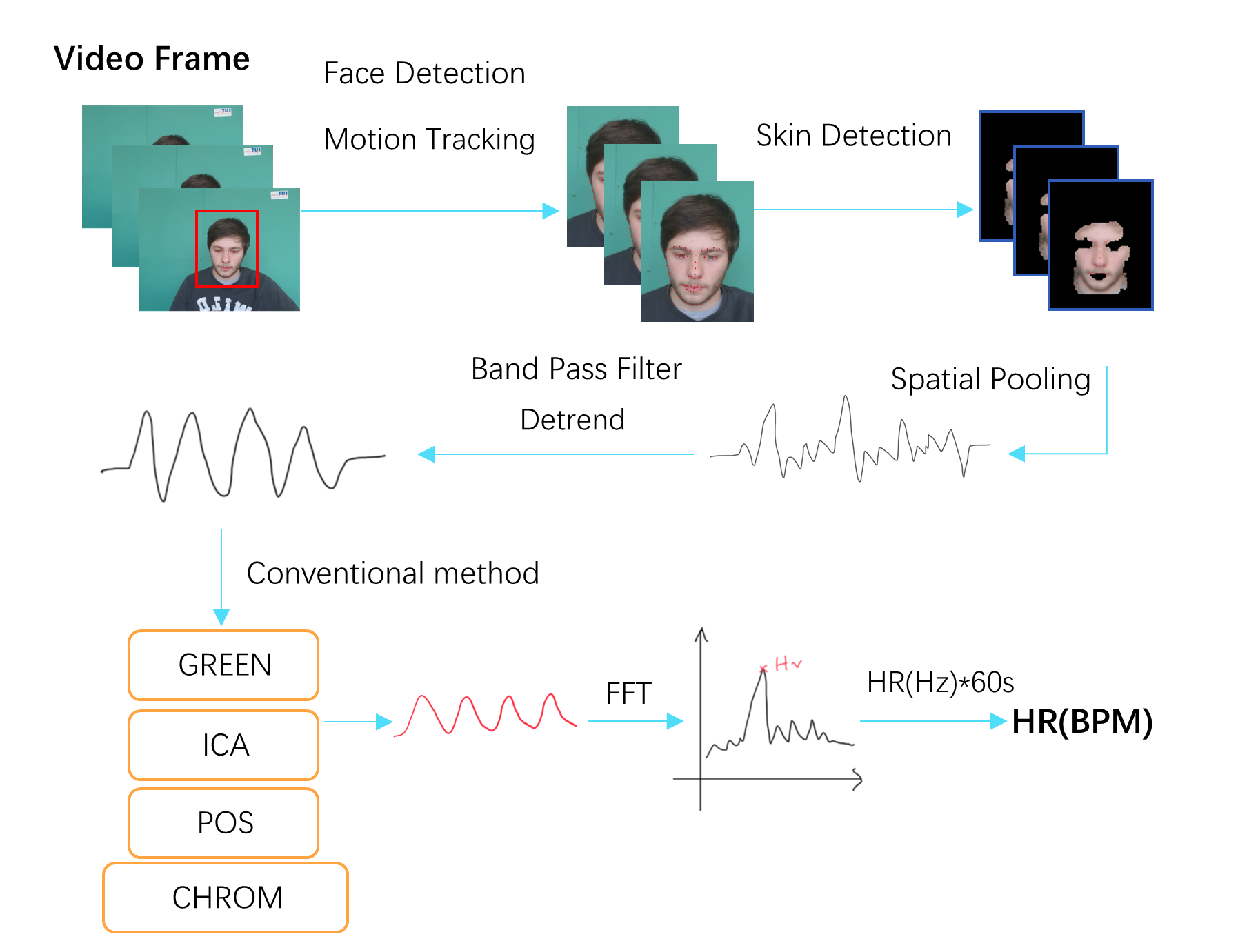}
	\caption{Traditional method workflow}
	\label{traditionalworkflow}
\end{figure}

We compared four representative methods: GREEN\cite{verkruysseRemotePlethysmographicImaging2008}, CHROM\cite{6523142}, POS\cite{wangAlgorithmicPrinciplesRemote2016} and ICA\cite{NoncontactAutomatedCardiac}. For implementation, we used the open source toolbox iPhys\cite{iphys}. The basic workflow of the traditional method is shown in Fig. \ref{traditionalworkflow}.

First, we detect and track the bounding box of the face using the KLT algorithm~\cite{KLT}. Then, the skin area is detected, and the eyes and mouth are removed since they often bring the noise for non-rigid movements during blink and speech. Next, the pulse signal is extracted by spatial pooling and all mean values for each frame are concatenated as raw pulse trace $\mathbf{C}(\mathbf{i})=\left[\vec{C}^{R}(i), \vec{C}^{G}(i), \vec{C}^{B}(i)\right]^{T}$ for $i$ = 1, ..., $L$, where $L$ is the number of frames in the video. After that, the varying part induced by heart rate can be obtained by a bandpass filter and detrending. Finally, we apply different methods to raw pulse trace, transform the signal into the frequency domain using Fast Fourious Transform (FFT), and find the peaks to estimate the heart rate.

\subsubsection{Deep learning-based methods}
We evaluate the performance of three typical deep learning-based methods: Deepphys \cite{chenDeepphysVideobasedPhysiological2018}, rPPGNet \cite{yuRemoteHeartRate2019}, and Physnet \cite{yuRemotePhotoplethysmographSignal2019}. The Deepphys is a two-dimensional CNN-based network that uses an attention mechanism to learn the skin map. The rPPGNet is a three-dimensional CNN-based network that uses soft attention to make the model focus on the skin area. The Physnet uses temporal encoder-decoder structure for rPPG task, which is applied in the action recognition task. 
% We employ the three dimensional CNN-based Physnet in \cite{yuRemotePhotoplethysmographSignal2019} as tested method.
The basic procedure of deep learning-based method can be formulated as below.
\begin{equation}
  \left[\hat{y}_{1}, \hat{y}_{2}, \ldots, \hat{y}_{T}\right]=\Omega\left(\Phi\left(\left[x_{1}, x_{2}, \ldots, x_{T}\right] ; \theta\right) ; w\right)
  \end{equation}

\noindent where $\left[y_{1}, y_{2}, \ldots, y_{T}\right]$ are the ground truths collected by finger oximeter, $\left[x_{1}, x_{2}, \ldots, x_{T}\right]$ are the frames sampled from original video. In Deepphys, T is set to 2, meaning that two consecutive frames are used to compute normalized difference frame and outputs $\hat{y}$. 
First, a CNN backbone $\Phi$ extracts spatial-temporal feature with parameter $\theta$, then $\Omega$ is used for channel aggregation with parameter $w$. The estimated PPG is $\left[\hat{y}_{1}, \hat{y}_{2}, \ldots, \hat{y}_{T}\right]$.  
Deepphys uses 2D CNN as $\Phi$ to extract spatial information and uses soft attention to assign different weights on skin regions.
Physnet and rPPGNet use 3D CNN as $\Phi$ to model temporal signal and take into account the correlation between ground truth and output.
We have re-implemented the Deepphys algorithm since the author did not release the source code, for rPPGnet and Physnet, we directly used the open-source model.

\subsection{Experimental protocols}
On the one hand, we would like to compare the performance of different deep learning methods trained with the same protocol, i.e., trained and evaluated using the same dataset. On the other hand, our goal is to evaluate the performance of deep learning-based method under different lighting conditions. We provide a comprehensive performance comparison between different methods. We also explore potential approaches for increasing the performance of deep learning-based methods.

\subsubsection{Performance comparison under the same training protocol}
We utilize the UBFC-rPPG dataset to train and test different deep learning-based methods. Specifically,
we randomly divide 42 videos into training-set (37 videos) and test-set (5 videos). Since each video corresponds to one subject, the task is subject-independent.

For traditional methods, we only evaluated performance on the test-set for a fair comparison with the deep learning-based method.

For Deepphys\cite{chenDeepphysVideobasedPhysiological2018}, we reproduced the model and trained with the same learning rate and batch size. For Physnet\cite{yuRemotePhotoplethysmographSignal2019} and rPPGnet\cite{yuRemoteHeartRate2019}, we adopted the frame length of 128, 64 as single clip which are sampled from original video respectively.

\subsubsection{Performance comparison under different illuminations}

We evaluated the traditional methods and the deep learning-based methods trained with UBFC-rPPG dataset. For traditional methods, we followed the settings of iPhys\cite{iphys}, except that the skin pixel value range changed according to different lighting conditions. Therefore, the frequency range of the raw signal between 0.7 to 2.5 Hz was extracted. For deep learning-based methods, we used the best model trained by the protocol mentioned in Section II to cross-test the BH-rPPG dataset. We used the average HR for the evaluation protocol of traditional methods.

\subsection{Potential method for performance improvement}
Additionally, we investigate various strategies for enhancing the performance of deep learning-based methods.
For models to better predict HR in videos with low lighting conditions, a natural way is to enhance each dark video frame by illumination compensation techniques to make the frame visually more transparent. Another direction for improving model accuracy is data augmentation, which is ubiquitous in image classification when labeled data is scarce. However, the lighting variations augmentation remain under-explored. Therefore, we tackle this problem in two stages:(1) Applying image enhancement after training and (2) Generating more lighting variations data during training. 
We will describe the details of implementation as below.
\subsubsection{Image enhancement methods}
We use the following three typical image enhancement algorithms:
a) Histogram Equalization (HE), b) Gamma Correction (GC) and c) Zero-Reference Deep Curve Estimation (ZERO-DCE\cite{guo2020zero}). The HE is a common approach to make the frame more contrast. Since the complex illumination leads lighting distribution on the face to be imbalanced. In this case, we apply the HE to V channel in HSV color space, which is more related to brightness. Then we transform back to RGB color space for model inference. The GC is a traditional image enhancement technique to improve image quality. For videos of low-light conditions, we set gamma values as 2.5. Furthermore, we set gamma values as 0.8 for videos with high lighting intensity. ZERO-DCE\cite{guo2020zero} is a more recent low-light image enhancement method through estimate pixel-wise and high-order curves for dynamic range adjustment of a given image. We directly use the pre-training model to further enhance our videos in low-light conditions.

\subsubsection{Video data augmentations}
We apply the same brightness data augmentation for a single clip rather than individually applying brightness augmentation on each frame since random transformation for a single frame may break the original pixel distribution. 
For brightness augmentation, we modified the function provided in PyTorch\cite{paszke2019pytorch} and randomly changed the brightness parameter in the range of [0.5, 1.5].

\subsection{Performance evaluation metrics}
We used four evaluation metrics: the mean absolute error (MAE), the root mean square error (RMSE), Signal-to-Noise Ratio (SNR), and Bland-Altman plot~\cite{liRemoteHeartRate2014},\cite{tulyakovSelfadaptiveMatrixCompletion2016}.

\subsubsection{Mean absolute error}
\begin{equation}
  \mathrm{HR}_{\text {mae}}=\frac{1}{n} \sum_{i=1}^{n}\left|\mathrm{HR}_{\text {predict }}^{(i)}-\mathrm{HR}_{\text {gt }}^{(i)}\right|,
\end{equation}

\noindent where $n$ is the total number of samples, $\mathrm{HR}_{\text {predict }}^{(i)}$ is the HR estimated by rPPG from $i$th samples, $\mathrm{HR}_{\text {gt }}^{(i)}$ is the ground-truth HR from $i$th samples.
  \subsubsection{Root mean square error}
  \begin{equation}
  \mathrm{HR}_{r m s e}=\sqrt{\frac{1}{n} \sum_{i=1}^{n}\left(\mathrm{HR}_{\mathrm{predict}}^{(i)}-\mathrm{HR}_{\text {gt }}^{(i)}\right)^{2}}.
  \end{equation}

\subsubsection{Signal-to-noise ratio}
The SNR computes the ratio of the energy around the fundamental frequency plus the first harmonic of the pulse signal and the remaining energy contained in the spectrum.
\begin{equation}
  \mathrm{SNR}=10 \log _{10}\left(\frac{\sum_{f=30}^{240}\left(U_{t}(f) \hat{S}(f)\right)^{2}}{\left.\sum_{f=30}^{240}\left(1-U_{t}(f)\right) \hat{S}(f)\right)^{2}}\right).
\end{equation}

Here we followed the same definition in \cite{6523142}, where $\hat{S}(f)$ is the spectrum of the pulse signal, $S, f$ is the frequency in beats per minute, and $U_{t}(f)$ is a binary template window.

\subsubsection{Bland-Altman plot} This plot demonstrates the consistency between two signals. The differences between the heart rate estimated by rPPG algorithm and the ground truth are plotted against the system average. We show the mean, standard deviation (SD), 95\% agreement limits (±1.96SD) of the difference.

\section{Experimental Results}\label{sec:experiment results}
% The performances on public dataset are reported in Table \ref{ubfcresults}. 
% Fig. \ref{fig:differentlight} and Fig. \ref{bland-altman} show the experimental results in different lighting conditions. The numerical values are shown in Table \ref{differentlightingtable}. 
\subsection{Performance comparison under the same protocol}
Table \ref{ubfcresults} demonstrates the performance of traditional methods and deep learning-based methods. Note that the deep learning-based methods are trained and tested under the same protocol using the UBFC-rPPG dataset. The Physnet\cite{yuRemotePhotoplethysmographSignal2019} achieves the best within UBFC-rPPG. In comparison to the traditional methods, the deep learning-based methods perform admirably. These results suggest that deep learning-based methods indeed demonstrate superior performance.
\begin{table}[h]
\centering
\caption{The results of UBFC-rPPG. The \textbf{bold} font means the best performance across all methods.}
\label{ubfcresults}
\begin{center}
\begin{tabular}{lccc} 
\hline
                               & Method   & MAE$~\downarrow$   & RMSE$~\downarrow$   \\ 
\hline
\multirow{3}{*}{Deep learning method} & Deepphys\cite{chenDeepphysVideobasedPhysiological2018} & 3.71  & 5.27   \\
                               & rPPGNet\cite{yuRemoteHeartRate2019}  & 3.24  & 4.97   \\
                               & Physnet\cite{yuRemotePhotoplethysmographSignal2019}  & \textbf{2.33}  & \textbf{3.04}   \\ 
\hdashline[1pt/1pt]
\multirow{4}{*}{Traditional method}  & CHROM\cite{6523142}    & 9.13  & 15.00  \\
                               & GREEN\cite{verkruysseRemotePlethysmographicImaging2008}    & 20.90 & 29.33  \\
                               & ICA\cite{NoncontactAutomatedCardiac}      & 8.62  & 13.54  \\
                               & POS\cite{wangAlgorithmicPrinciplesRemote2016}      & 15.39 & 27.22  \\
\hline
\end{tabular}
\end{center}
\end{table}

\subsection{Performance comparison under different illumination}
\begin{figure}[!t]
    \subfloat{%
    
    \includegraphics[clip, width= 0.4\textwidth]{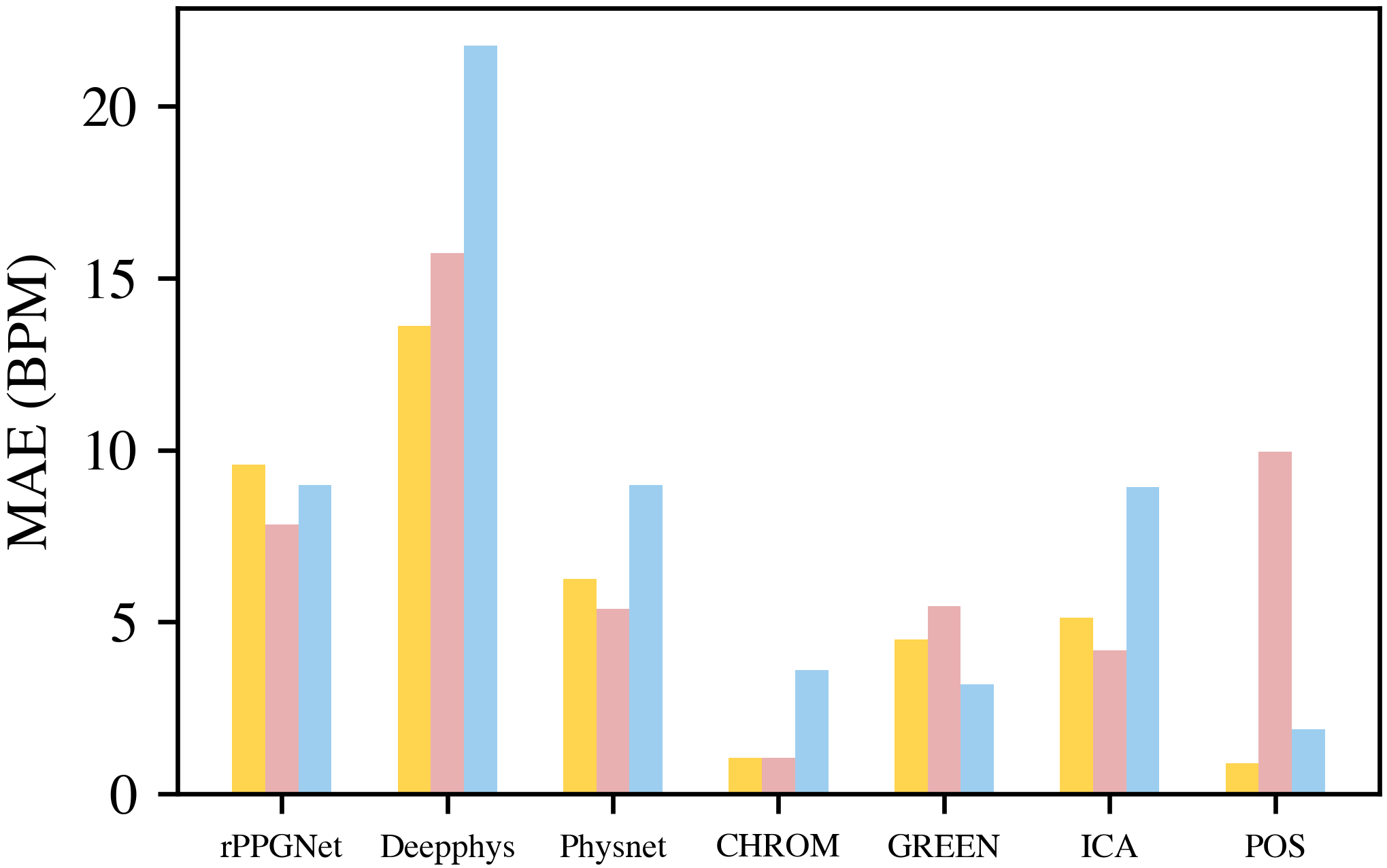}}
    
    \subfloat{%
    \includegraphics[clip, width= 0.4\textwidth]{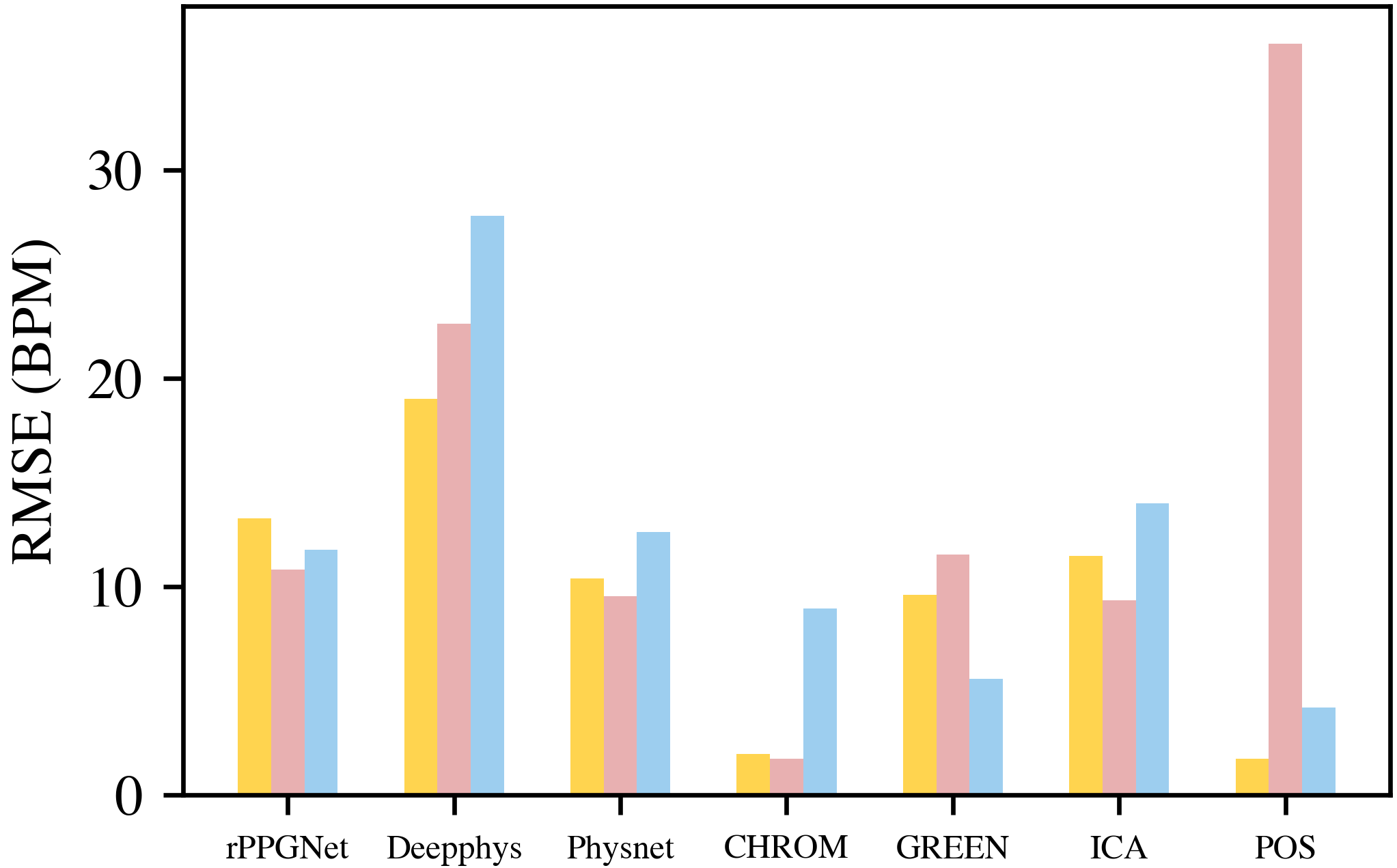}}
  
    \subfloat{%
    \includegraphics[clip, width= 0.4\textwidth]{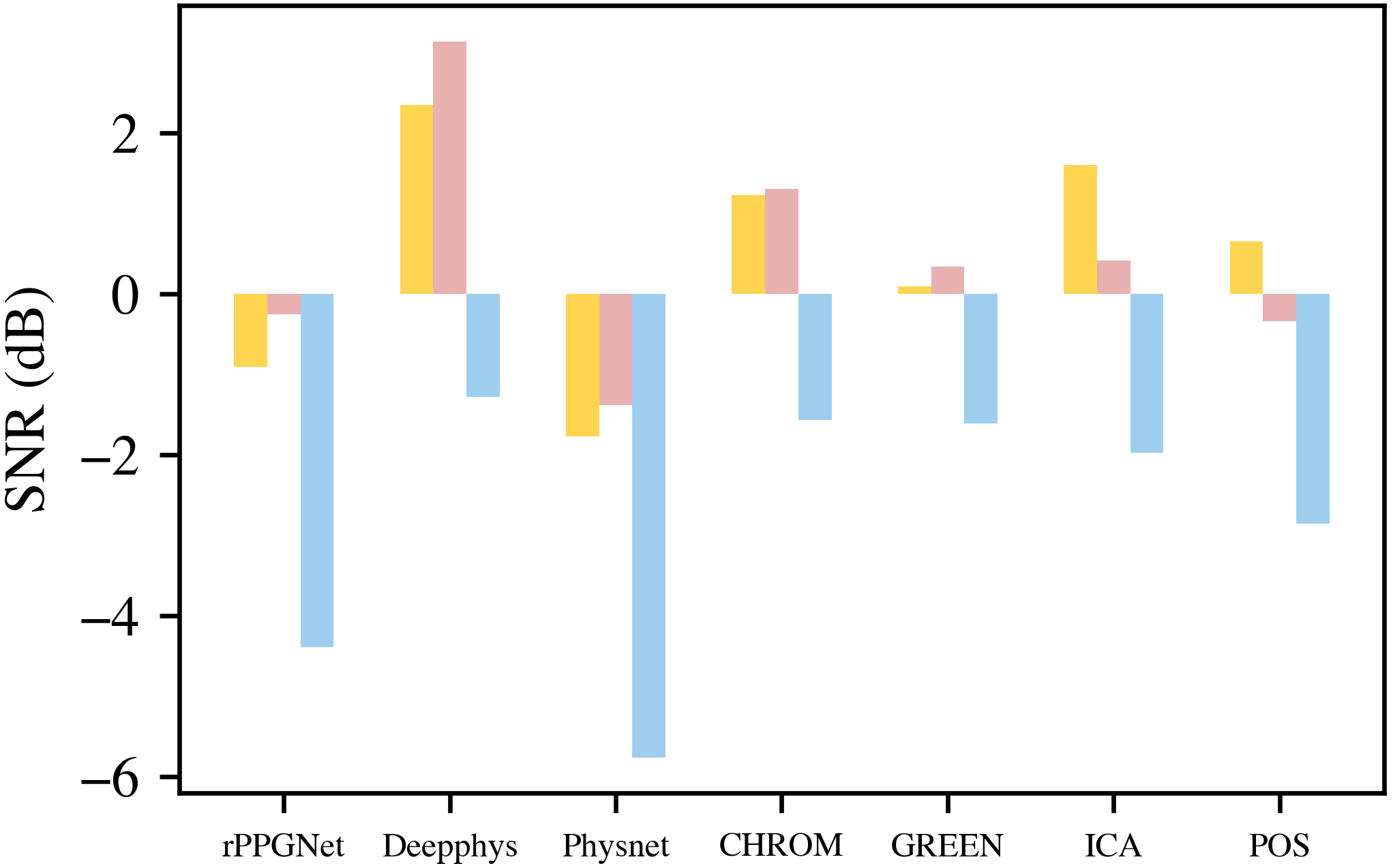}}
    \caption{Performance comparison between different methods. The three subplots shows the MAE, RMSE, SNR, respectively. The x-axis represents different methods, different colors shows the three lighting conditions: high, medium and low.}
    \label{fig:differentlight}
  \end{figure}

Table \ref{differentlightingtable} shows the performance of traditional methods and deep learning-based methods under different illuminations. From the MAE plot in Fig. \ref{fig:differentlight}, we observed that the CHROM in medium lighting conditions achieved the best results among all methods at 1.04 BPM. And the medium-light and high-light show similar good results. 
The second best method is performed by POS in high-light. Besides the result in medium-light, which shows 9.96 BPM in MAE, the performance achieved by POS is under 2 BPM.  
As for deep learning-based method,  Physnet reached an MAE of 5.37 BPM in medium-light, which is the best performance among deep learning approaches. However, it still falls behind most of the traditional methods, except for ICA of low-light and POS of medium-light, which performs poorly in medium and low lighting conditions, respectively. The results reveal that traditional methods are more robust to different illumination scenarios. 
In the RMSE plots, the lowest performance in terms of RMSE is achieved by CHROM, which outperforms the best model for deep learning-based $80\%$.  
% Both rPPGNet and Physnet perform well in medium lighting conditions and Deepphys produce good results in low lighting intensity in three distinct lighting settings. 

%  图空白

\begin{table}[htb]
\centering
\caption{The results of different lighting conditions. The DL shows the deep learning-based methods and TR represents the traditional methods.The best performance between three lighting conditions is in \textbf{bold}.The best performance of all methods is in \underline{underlined}.}
\label{differentlightingtable}
\begin{tabular}{lccccc} 
\hline
                            & \multicolumn{1}{l}{Methods} & \multicolumn{1}{l}{Illum} & \multicolumn{1}{l}{MAE$~\downarrow$} & \multicolumn{1}{l}{RMSE$~\downarrow$} & \multicolumn{1}{l}{SNR$~\uparrow$}  \\ 
\hline
\multirow{9}{*}{DL}  & \multirow{3}{*}{rPPGNet\cite{yuRemoteHeartRate2019}}    & low                                     & 8.99                    & 11.77                     & -4.39                   \\
                            &                             & medium                                  & \textbf{7.82}                    & \textbf{10.83}                     & \textbf{-0.26}                    \\
                            &                             & high                                    & 9.58                    & 13.29                     & -0.90                    \\ 
\cline{2-6}
                            & \multirow{3}{*}{Deepphys\cite{chenDeepphysVideobasedPhysiological2018}}   & low                                     & 21.76                   & 27.82                    & -1.28                    \\
                            &                             & medium                                  & 15.73                   & 22.61                    & \textbf{3.14}                     \\
                            &                             & high                                    & \textbf{13.60}                   & \textbf{19.01}                    & 2.35                    \\ 
\cline{2-6}
                            & \multirow{3}{*}{Physnet\cite{yuRemotePhotoplethysmographSignal2019}}    & low                                     & 8.98                   & 12.61                    & -5.76                    \\
                            &                             & medium                                  & \textbf{5.37}                    & \textbf{9.55}                    & \textbf{-1.38}                    \\
                            &                             & high                                    & 6.25                   & 10.41                    & -1.77                    \\ 
\hdashline[1pt/1pt]
\multirow{12}{*}{TR} & \multirow{3}{*}{CHROM\cite{6523142}}      & low                                     & 3.6                    & 8.95                     & -1.57                    \\
                            &                             & medium                                  & \underline{\textbf{1.04}}                    & \underline{\textbf{1.73}}                     & \underline{\textbf{1.30}}                     \\
                            &                             & high                                    & 1.06                    & 1.96                     & 1.23                     \\ 
\cline{2-6}
                            & \multirow{3}{*}{GREEN\cite{verkruysseRemotePlethysmographicImaging2008}}      & low                                     & \textbf{3.19}                    & \textbf{5.58}                     & -1.61                    \\
                            &                             & medium                                  & 5.46                    & 11.54                     & \textbf{0.34}                     \\
                            &                             & high                                    & 4.50                    & 9.61                     & 0.09                     \\ 
\cline{2-6}
                            & \multirow{3}{*}{ICA\cite{NoncontactAutomatedCardiac}}        & low                                     & 8.93                    & 14.02                    & -1.98                  \\
                            &                             & medium                                  & \textbf{4.17}                    & \textbf{9.35}                     & 0.42                     \\
                            &                             & high                                    & 5.13                    & 11.49                     & \textbf{1.60}                     \\ 
\cline{2-6}
                            & \multirow{3}{*}{POS\cite{wangAlgorithmicPrinciplesRemote2016}}        & low                                     & 1.88                    & 4.21                     & -2.85                    \\
                            &                             & medium                                  & 9.96                    & 36.07                     & -0.34                     \\
                            &                             & high                                    & \textbf{0.89}                    & \textbf{1.73}                     & \textbf{0.65}                     \\
\hline
\end{tabular}
\end{table}

From the SNR plot in Fig. \ref{fig:differentlight}, we found that the conventional methods achieved positive values in high and medium lighting conditions. However, only the Deepphys achieve positive SNR in medium and high light conditions. It is notable that deep learning-based methods are significantly inferior to traditional methods in terms of generalization ability, and different algorithms perform inconsistently in different illuminations. Deepphys works better in high-light conditions. Both rPPGNet and Physnet are more effective in medium-light conditions. This is possibly because that the medium-light is more similar to UBFC-rPPG.
Fig. \ref{bland-altman} depicts the ground-truth HR and the estimated HR under three different lighting situations, which is to illustrates the estimation consistency of various methods, in which each scatter point represents the estimation error of HR. Traditional methods are much more consistent with ground-truth HR than deep learning-based methods, i.e., the distribution of scatter points is very close to the x-axis. Both of rPPGNet and Deepphys have errors at above 30 BPM, indicating the fails of HR prediction. As for Physnet, the errors in low lighting conditions stabilize at a low level within the range of [-10, 10]; only a small part of outlier appears at around 20 BPM. However, the traditional methods have more consistent results.
\newlength{\tempheight}
\newlength{\tempwidth}
  
  \newcommand{\rowname}[1]% #1 = text
  {\rotatebox{90}{\makebox[\tempheight][c]{\textbf{#1}}}}
  
  \newcommand{\columnname}[1]% #1 = text
  {\makebox[\tempwidth][c]{\textbf{#1}}}
  \begin{figure*} 
    \setlength{\tempwidth}{.3\linewidth}
  \settoheight{\tempheight}{\includegraphics[width=\tempwidth]{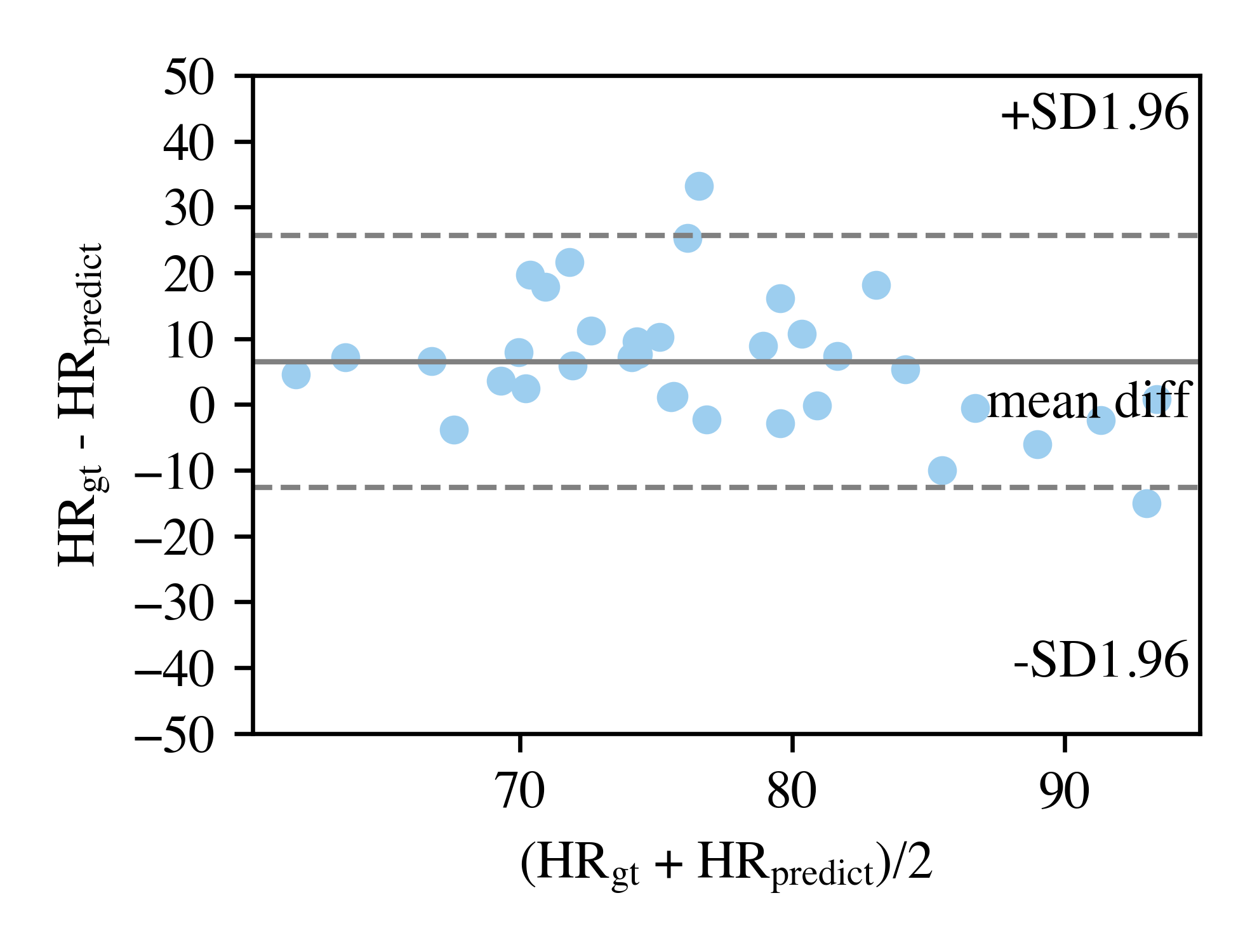}}%
  \centering
  \hspace{\baselineskip}
  \columnname{Low lighting}\hfil
  \columnname{Medium lighting}\hfil
  \columnname{High lighting}\\
  \rowname{rPPGNet}
      % \centering
    \subfloat{%
         \includegraphics[width=\tempwidth]{IMAGES/hms_bland__rppg_PeakDetection_lowlamp.png}}
      \hfil
    \subfloat{%
          \includegraphics[width=\tempwidth]{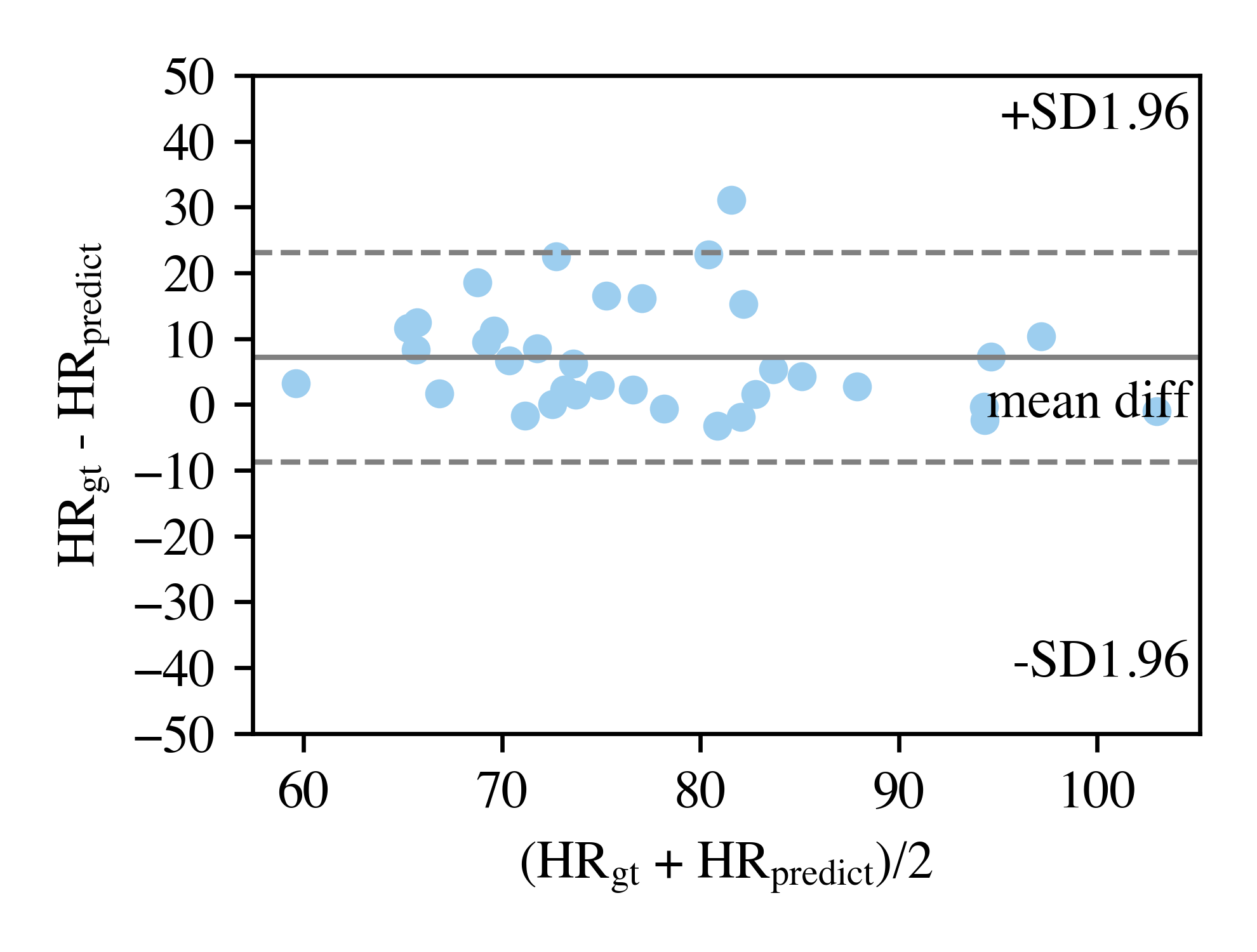}}
      \hfil
    \subfloat{%
      \includegraphics[width=\tempwidth]{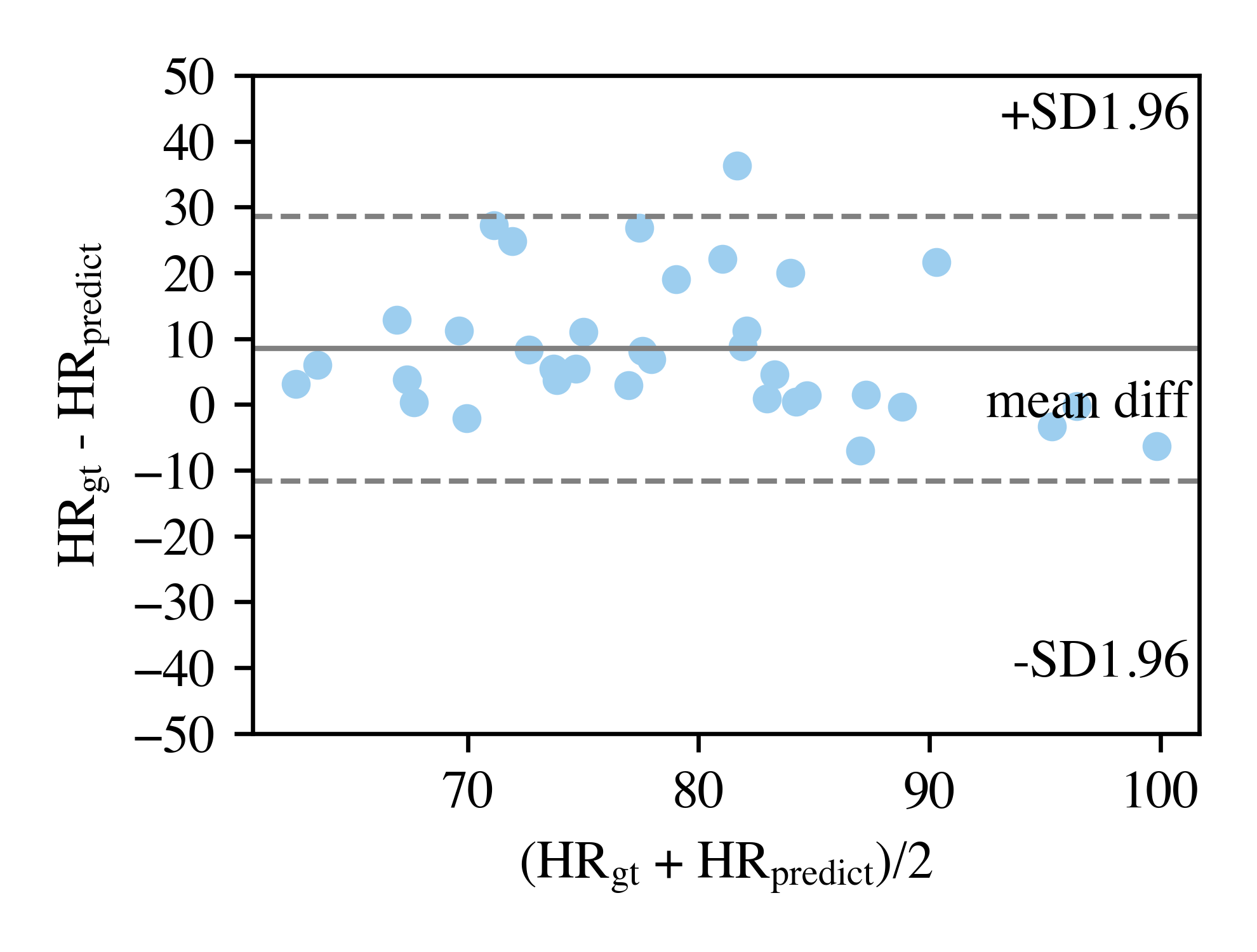}}
      \\
  \rowname{Deepphys}
    \subfloat{%
         \includegraphics[width=\tempwidth]{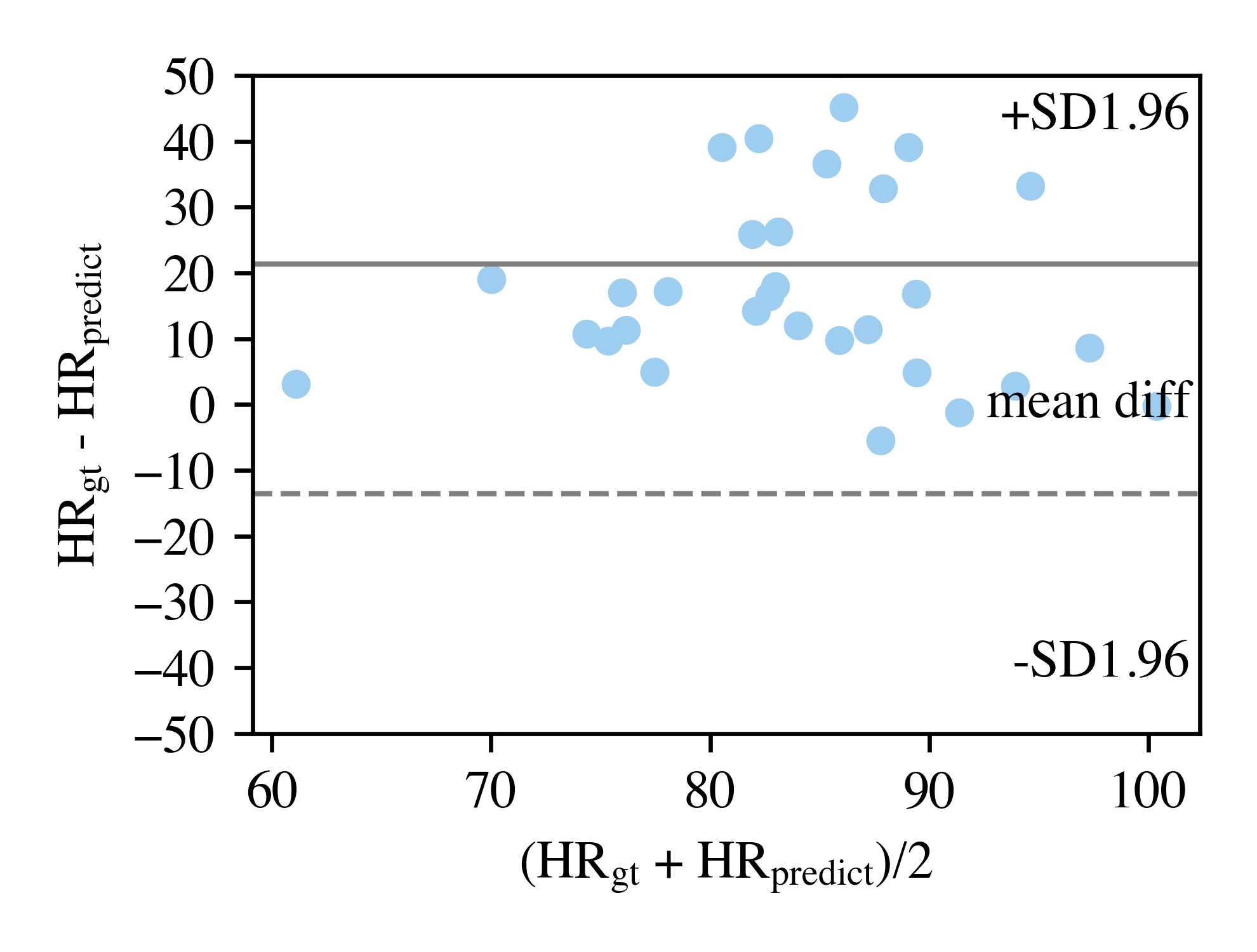}}
      \hfil
    \subfloat{%
          \includegraphics[width=\tempwidth]{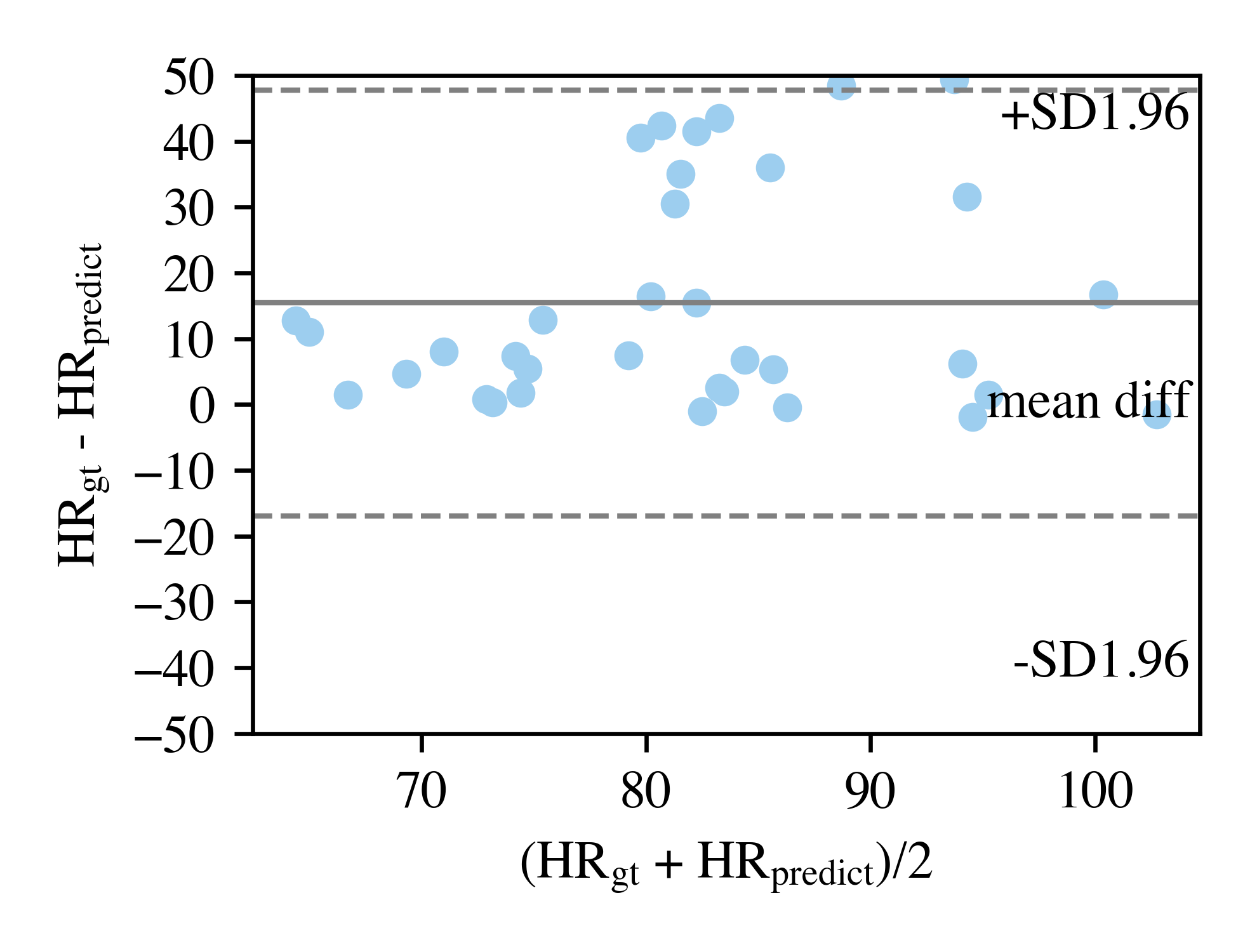}}
      \hfil
    \subfloat{%
      \includegraphics[width=\tempwidth]{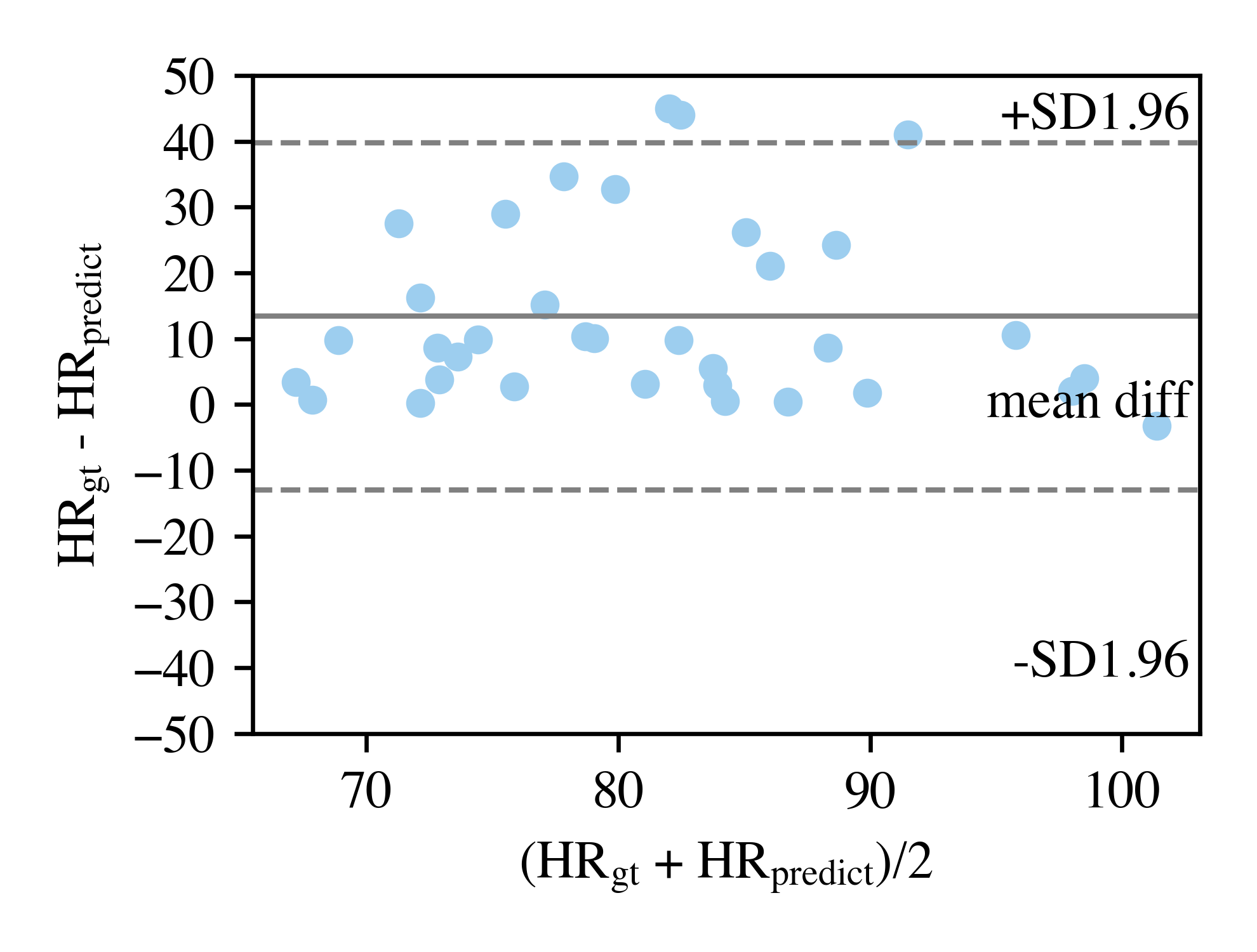}}
      \\
  \rowname{Physnet}
    \subfloat{%
         \includegraphics[width=\tempwidth]{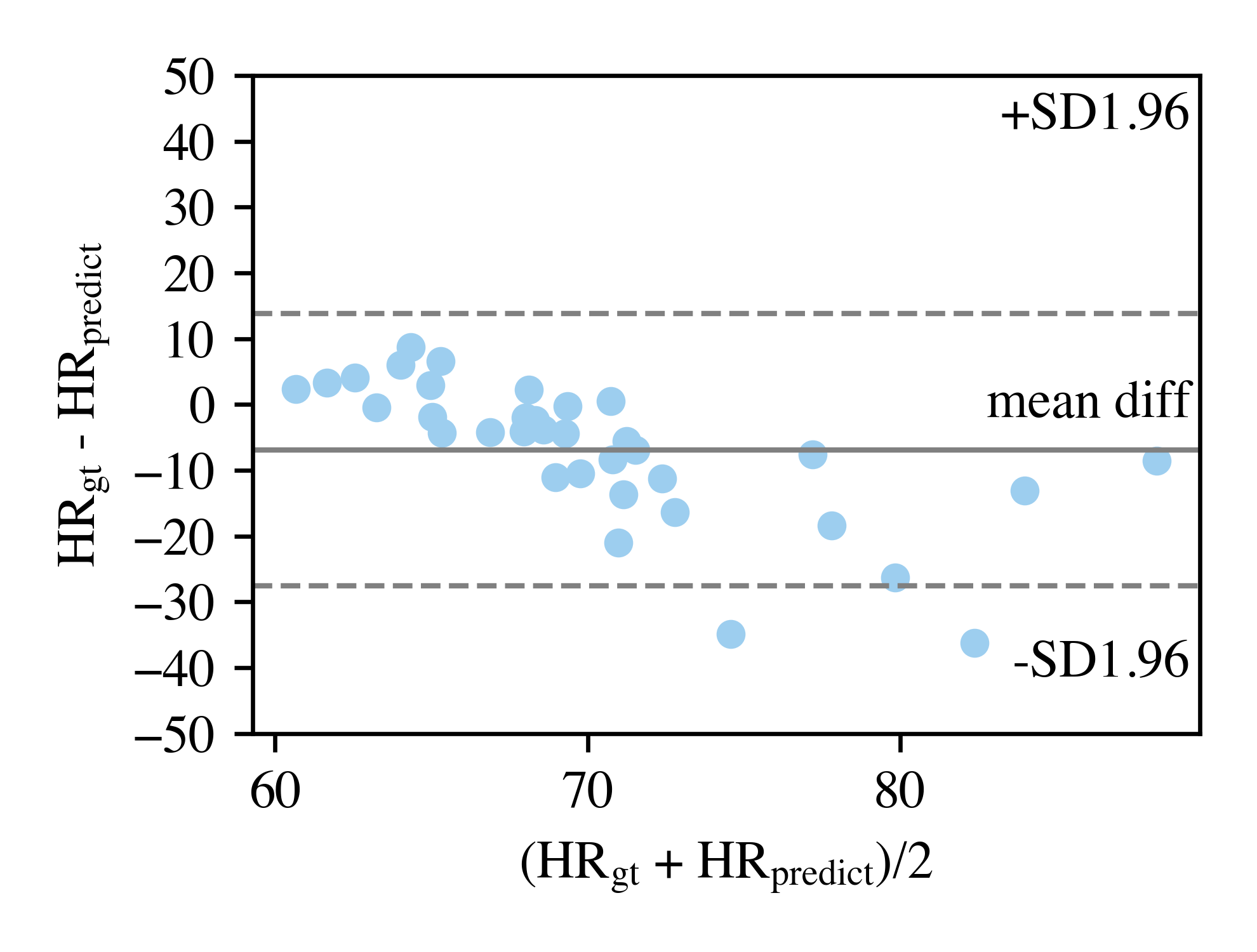}}
      \hfil
    \subfloat{%
          \includegraphics[width=\tempwidth]{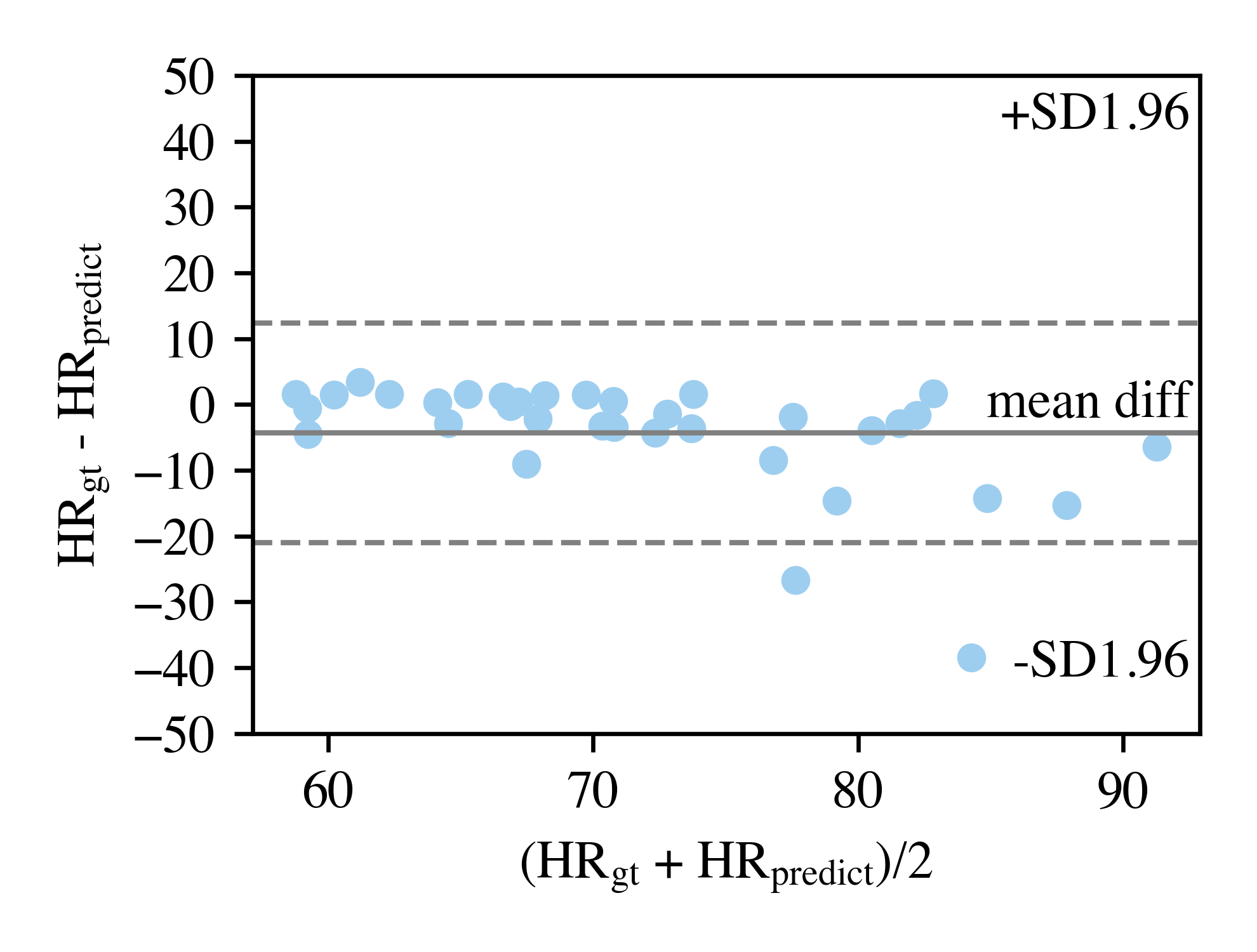}}
      \hfil
    \subfloat{%
      \includegraphics[width=\tempwidth]{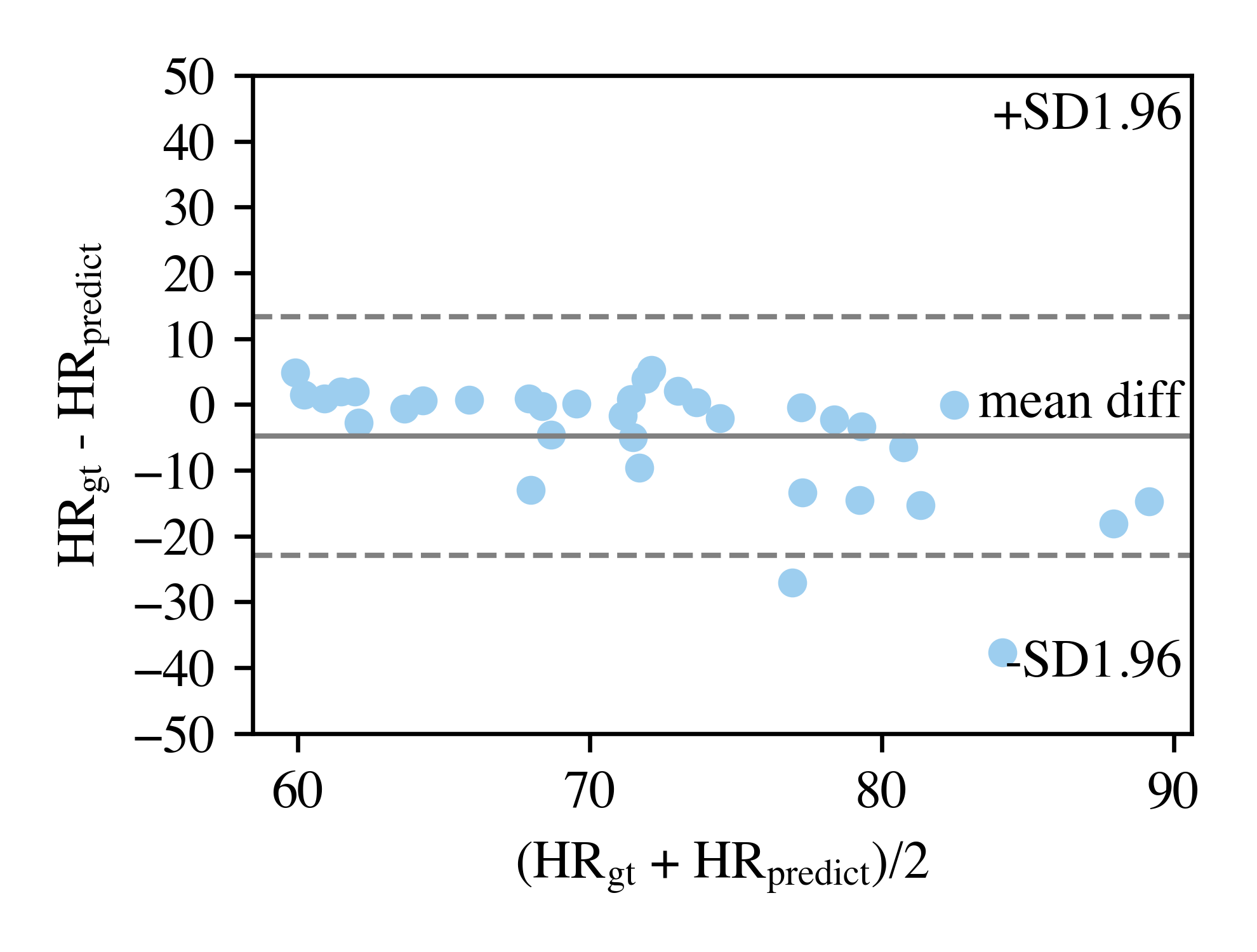}}
      \\
  \rowname{CHROM}
    \subfloat{%
         \includegraphics[width=\tempwidth]{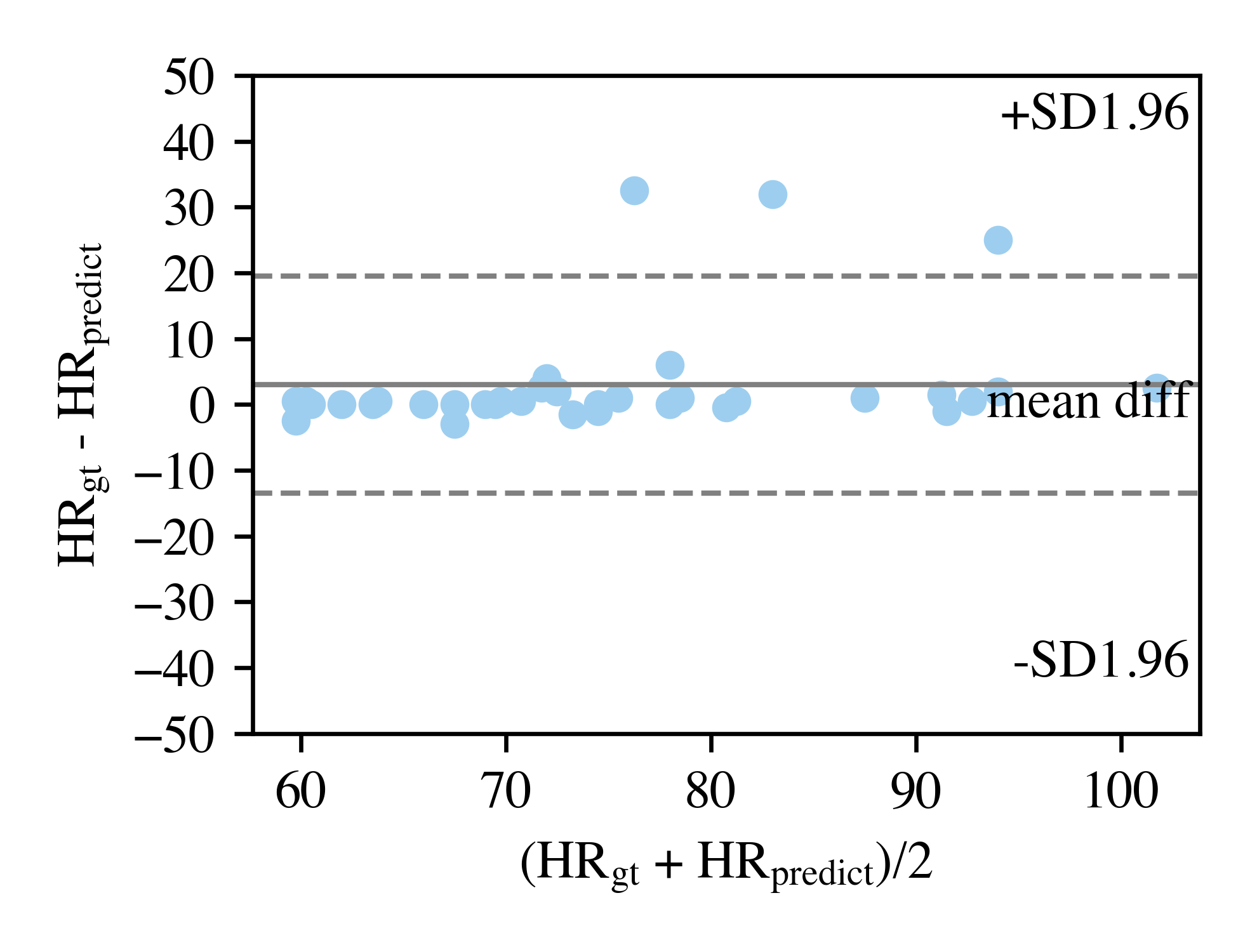}}
      \hfil
    \subfloat{%
          \includegraphics[width=\tempwidth]{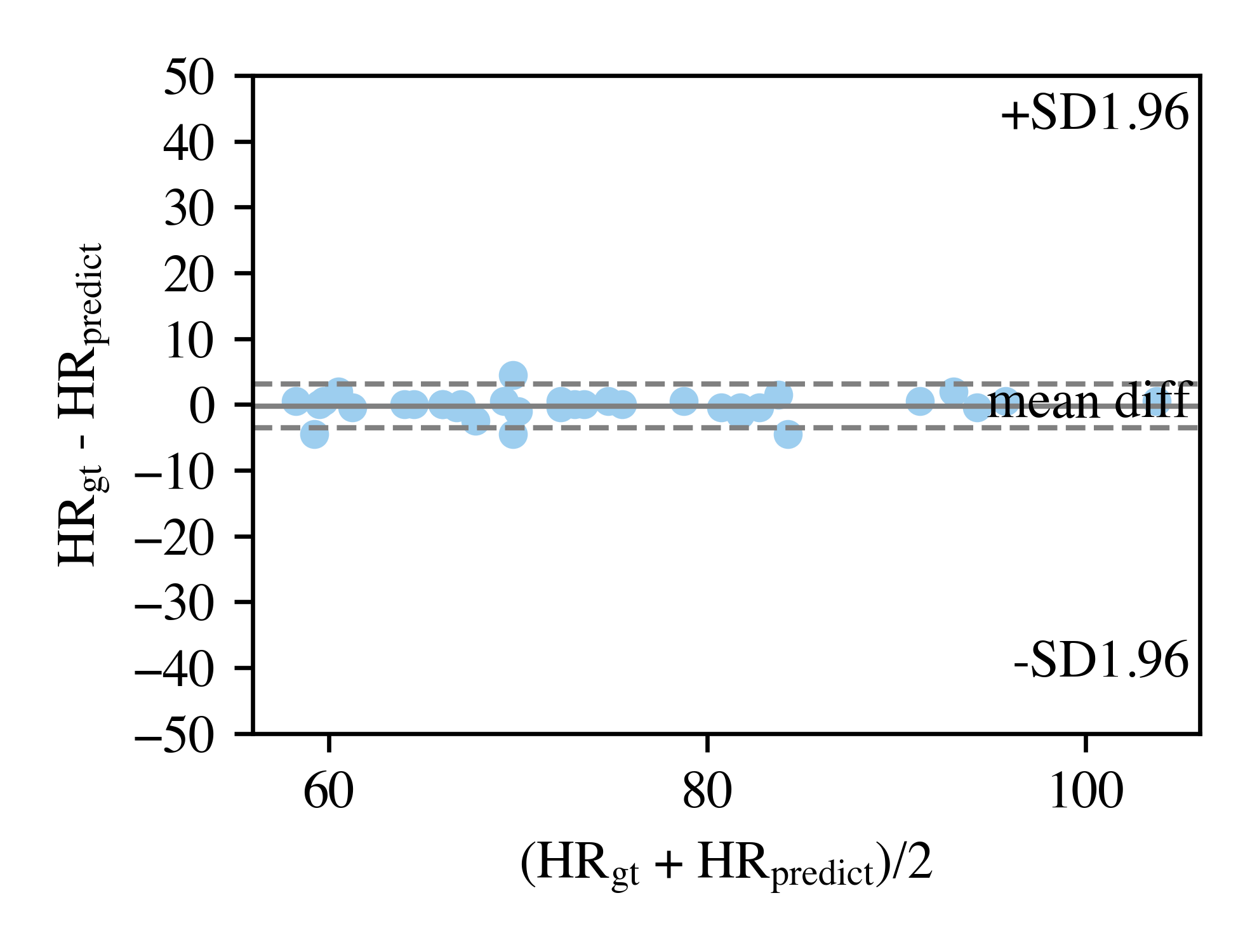}}
      \hfil
    \subfloat{%
      \includegraphics[width=\tempwidth]{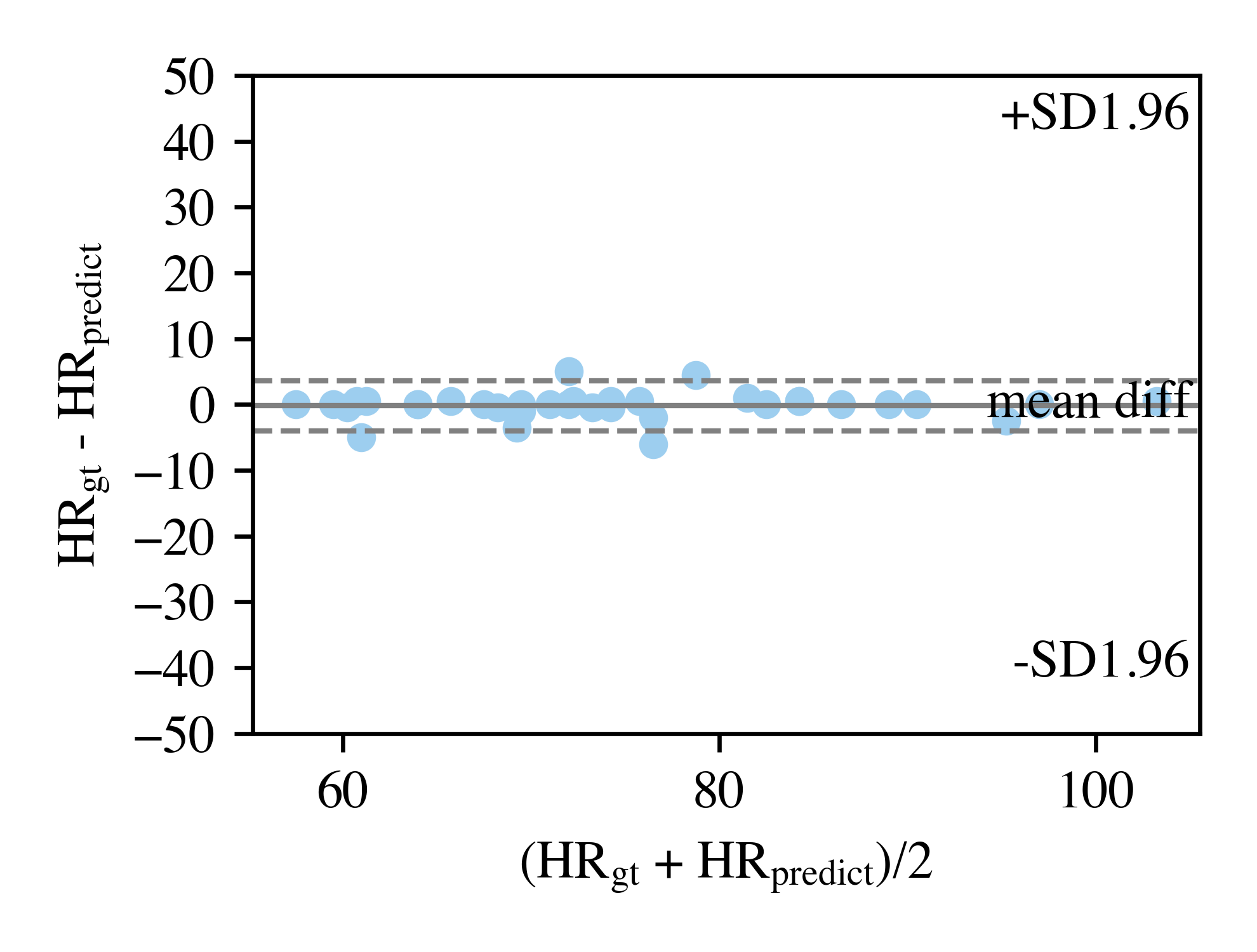}}
      \\
    \caption{The Bland-Altman plots of the rPPGNet, Deepphys, Physnet, and CHROM methods under different light conditions. Each row shows different methods, while each column shows different lighting conditions. The Bland-Altman plots illustrate the degree of agreement between the estimated HR and ground-truth HR. The solid lines represent the mean value, while the dashed lines represent 95\% limits of agreement.}   
    \label{bland-altman} 
  \end{figure*}

\subsection{Results of applying image enhancement methods to BH-rPPG}
Table \ref{image_enhancement} demonstrates that HE method has a negative effect on all three lighting conditions videos, except for rPPGNet and Physnet, positive gain to SNR only. Similar to HE, Zero-DCE contributes little to the performance of improvement. One possible reason might be that temporal consistency is significant for HR estimation and illumination compensation techniques break the coherency of pixel variation between frames. 
As for GC-H, we see that Physnet achieves 9.83$\%$, 6.87$\%$ and 4.81$\%$ improvement in MAE, RMSE, SNR, respectively. It might be because the GC successfully aligns the brightness of BH-rPPG in high-light to UBFC-rPPG dataset. 
\begin{table}
  \centering
  \caption{Performance of deep learning-based methods on the BH-rPPG enhanced by GC, HE and ZERO-DCE. The low, mid and high represents the videos under low, mid and high lighting conditions.}
  \label{image_enhancement}
  \begin{tabular}{ccccc} 
  \hline
  \multicolumn{1}{l}{Data} & \multicolumn{1}{l}{Metric} & \multicolumn{1}{l}{rPPGNet} & \multicolumn{1}{l}{Deepphys} & \multicolumn{1}{l}{Physnet}  \\ 
  \hline
  \multirow{3}{*}{GC-L}     & MAE                             & 19.19                       & 24.00                        & 10.26                        \\
                               & RMSE                            & 21.95                       & 30.12                        & 13.16                        \\
                               & SNR                             & $-3.75_{\textcolor{red}{\blacktriangle14.63\%}}$                       & $-0.96_{\textcolor{red}{\blacktriangle25.10\%}}$                        & -5.85                        \\ 
  \hline
  \multirow{3}{*}{GC-H}    & MAE                             & 9.42                        & 14.48                        & $5.64_{\textcolor{red}{\blacktriangledown9.83\%}}$                         \\
                               & RMSE                            & 12.99                       & 19.97                        & $9.69_{\textcolor{red}{\blacktriangledown6.87\%}}$                         \\
                               & SNR                             & $-0.65_{\textcolor{red}{\blacktriangle28.14\%}}$                       & $-2.36_{\textcolor{red}{\blacktriangle0.58\%}}$                        & $-1.38_{\textcolor{red}{\blacktriangle4.81\%}}$                        \\ 
  \hline
  \multirow{3}{*}{HE-L}      & MAE                             & 16.71                       & 29.50                        & 10.19                        \\
                               & RMSE                            & 20.11                       & 33.04                        & 12.96                        \\
                               & SNR                             & $-1.97_{\textcolor{red}{\blacktriangle85.97\%}}$                       & -2.19                        & $-5.09_{\textcolor{red}{\blacktriangle11.59\%}}$                        \\ 
  \hline
  \multirow{3}{*}{HE-M}      & MAE                             & 9.95                        & 19.07                        & 6.67                         \\
                               & RMSE                            & 13.15                       & 26.53                        & 10.64                        \\
                               & SNR                             & -0.64                       & 1.84                         & -2.76                        \\ 
  \hline
  \multirow{3}{*}{HE-H}     & MAE                             & 13.17                       & 17.94                        & 7.43                         \\
                               & RMSE                            & 16.87                       & 24.75                        & ~11.11                       \\
                               & SNR                             & -2.24                       & 1.51                         & -3.45                        \\ 
  \hline
  \multirow{3}{*}{Zero-L}   & MAE                             & 22.16                       & 24.72                        & 10.59                        \\
                               & RMSE                            & 25.36                       & 30.52                        & 13.99                        \\
                               & SNR                             & -4.61                       & $-1.08_{\textcolor{red}{\blacktriangle15.19\%}}$                        & -6.78                        \\
  \hline
  \end{tabular}
  \end{table}
  \begin{table}
    \centering
    \caption{Performance of the model training with brightness augmentation on the BH-rPPG.Red font represents the percentage of performance improvement. L, M and H mean three lighting conditions.}
    \label{brightness}
    \begin{tabular}{ccccc} 
    \hline
    \multicolumn{1}{l}{Methods} & \multicolumn{1}{l}{Illum} & \multicolumn{1}{l}{MAE$~\downarrow$} & \multicolumn{1}{l}{RMSE$~\downarrow$} & \multicolumn{1}{l}{SNR$~\uparrow$}  \\ 
    \hline
    \multirow{3}{*}{rPPGNet}    & L                                     & 18.77                   & 21.09                    & -4.78                    \\
                                & M                                  & $\textbf{6.98}_{\textcolor{red}{\blacktriangledown10.87\%}}$                    & \textbf{10.87}                    & $\textbf{-0.18}_{\textcolor{red}{\blacktriangle30.05\%}}$                    \\
                                & H                                    & $9.45_{\textcolor{red}{\blacktriangledown1.39\%}}$                    & 13.46                    & $-1.32_{\textcolor{red}{\blacktriangle45.07\%}}$                    \\ 
    \hline
    \multirow{3}{*}{Deepphys}   & L                                  & $21.46_{\textcolor{red}{\blacktriangledown1.40\%}}$                   & $26.25_{\textcolor{red}{\blacktriangledown5.66\%}}$                    & -1.32                    \\
                                & M                                  & \textbf{16.88}                   & \textbf{23.82}                    & \textbf{0.50}                     \\
                                & H                                  & 19.75                   & 25.15                    & -0.95                     \\ 
    \hline
    \multirow{3}{*}{Physnet}    & L                                  & $7.98_{\textcolor{red}{\blacktriangledown11.15\%}}$                    & $12.24_{\textcolor{red}{\blacktriangledown2.93\%}}$                    & $-4.97_{\textcolor{red}{\blacktriangle13.62\%}}$                    \\
                                & M                                  & $4.44_{\textcolor{red}{\blacktriangledown17.38\%}}$                    & $7.26_{\textcolor{red}{\blacktriangledown23.95\%}}$                     & $\underline{\textbf{-0.48}}_{\textcolor{red}{\blacktriangle65.57\%}}$                    \\
                                & H                                  & $\underline{\textbf{3.18}}_{\textcolor{red}{\blacktriangledown49.19\%}}$                    & $\underline{\textbf{5.19}}_{\textcolor{red}{\blacktriangledown50.11\%}}$                     & $-0.91_{\textcolor{red}{\blacktriangle48.58\%}}$                   \\
    \hline
    \end{tabular}
    \end{table}
\subsection{Results of applying data augmentation to BH-rPPG}
In Table \ref{brightness}, the brightness jitter in training stage has great positive effect on rPPGNet and Physnet. 
% Physnet in high-light perform best results and outperform non-augmentation $49.19\%$ in MAE.
Compared with the non-augmentation approaches, the Physnet with data augmentation achieves the optimal performance in high-light condition, with an MAE gain of $49.19\%$, while the rPPGNet with data augmentation obtains an MAE improvement of $10.87\%$ and SNR improvement of $30.05\%$ in medium-light condition.
We notice that the Deepphys show poor improvements for data augmentation, partly because the 2D-CNN is hard to learn the illumination variation. 
Therefore, the results in Table \ref{brightness} exhibit that the brightness augmentation in the training stage is indeed more efficient than image enhancement methods to improve the performance. 

\section{Discussion}
\label{sec:discussion}
\subsection{Distribution of rPPG signal under uneven light}
\begin{figure}[htb]
    \centering
    \includegraphics[width=0.4\textwidth]{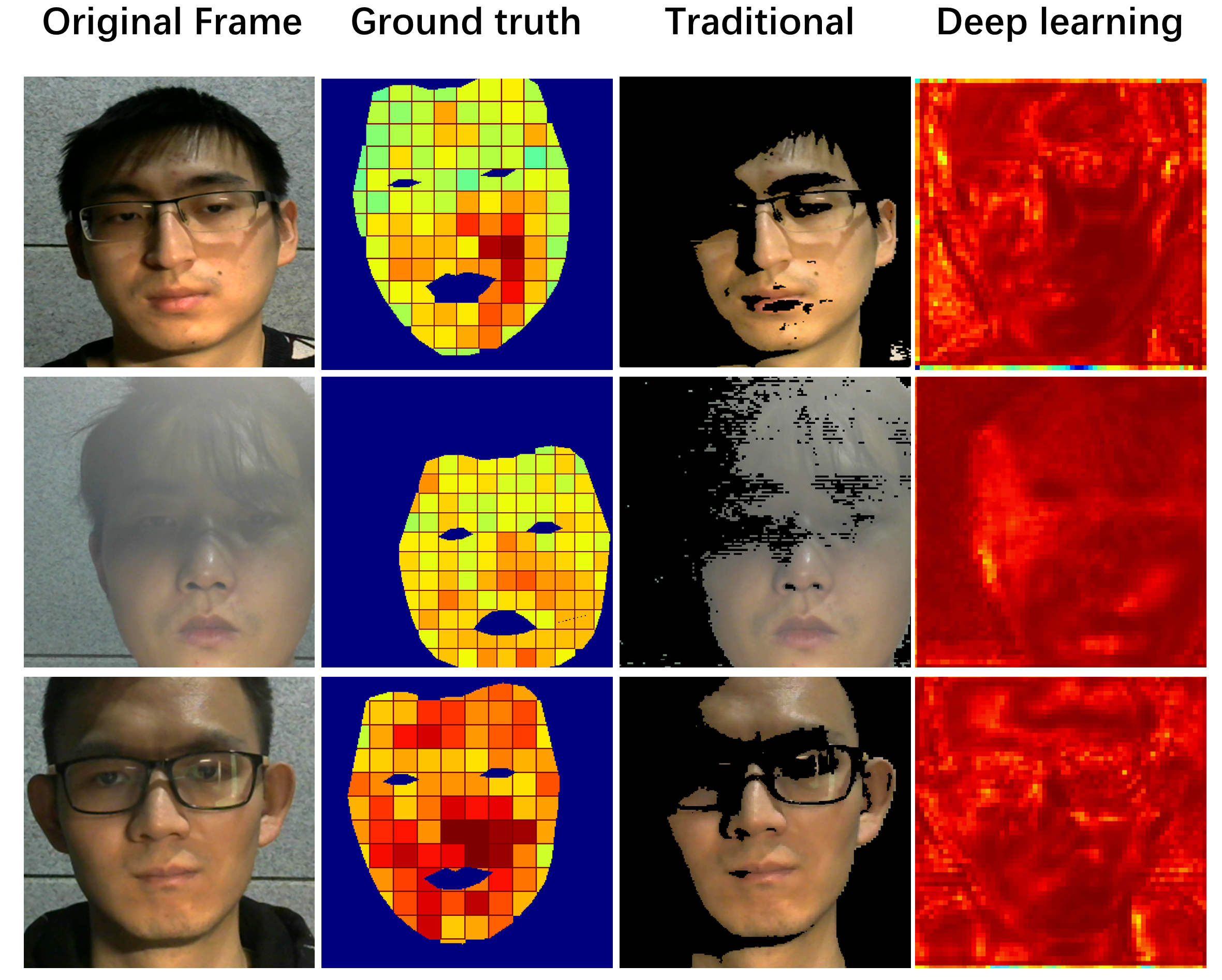}
    \caption{The visualization of ROI. The three columns show the results of the traditional method, the ground truth and the attention map extracted by rPPGNet.}
    \label{skinattention}
\end{figure}

To better understand the performance difference, we visualize the region-of-interest (ROI) in different methods. Fig. \ref{skinattention} shows the original frame, ground truth, preprocessing results of traditional method, and attention weights of intermediate steps in rPPGNet.

The POS method is selected as producing the ground truth map with the definition of SNR mentioned in Section III.E. Since the traditional method can accurately depict the actual distribution of the rPPG signal on the face. We can see from the original frame and ground truth that the brighter the region of the face, the higher the SNR of the rPPG signal. The varying light intensity changes the distribution of rPPG signal, which echoes the findings in \cite{wangAlgorithmicPrinciplesRemote2016} and \cite{wangRobustAutomaticRemote2017}.

However, the attention weights learned by deep learning methods demonstrate that the neural network focuses on background and skin area that is irrelevant to light. One possibility is that there is a domain gap between the training data (UBFC-rPPG dataset) and test data (BH-rPPG dataset). Because the skin tone and good lighting conditions in the training set of UBFC-rPPG is different from the test set of BH-rPPG. Lee \etal{}\cite{lee2020meta} proposed a meta-learning framework to update model weight, which may help model to adapt to different application situations. In addition, the skin branch in rPPGNet is a series of learnable weights optimized as the ground truth of binary skin mask which is generated by \cite{taylor2014adaptive}. When the lighting intensity, skin tone or environment changes, it is natural for the skin branch to produce the incorrect skin mask. We believe that finding the ROI branch is significant to deep learning-based rPPG task. While the varying lighting conditions have significant effect on the performance of deep learning-based methods.

In contrast to deep learning-based rPPG, traditional methods detect skin regions in the preprocessing steps which has been visualized in traditional columns of Fig.\ref{skinattention}. Although the lighting intensity distribution on face is uneven, the skin detection algorithm successfully locates the correct skin area that contains rPPG signal induced by cardiac activity. This may explain why traditional approaches perform better than deep learning-based methods under different illuminations. Furthermore, the average pooling for whole region can retrieve the information for brighter regions that contains more rPPG signal and smooth the darker regions where contribute less to rPPG. Additionally, due to the spatial redundancy of rPPG\cite{wangExploitingSpatialRedundancy2014}, some studies partition the face into distinct grids allows that each grid for different grids may be beneficial under varying lighting conditions.  
 
\subsection{Algorithm robustness to illumination variation}
In this paper, we investigated the performance of deep learning-based rPPG under varying lighting conditions and used conventional methods as a baseline. From Table \ref{ubfcresults} and Table \ref{differentlightingtable}, we found that deep learning-based methods perform well within UBFC-rPPG dataset and show poor performance in BH-rPPG dataset. Especially for Physnet, which shows the best performance within UBFC-rPPG, although perform the best results among deep learning-based methods, it still do not perform as accuracy as traditional method, such that CHROM. For this case, the appearance of background, subject's skin tone, and lighting conditions greatly impact the the deep learning-based method. Therefore, the ambient light needs to be chosen carefully.

As the conventional method, it performs poorly in the UBFC-rPPG dataset but well in BH-rPPG dataset. One possible reason for low accuracy in UBFC-rPPG is the motion effect on rPPG signal. The deep learning-based method learns the relationship between pixels values and HR by a large number of non-linear mapping, and the linear combination of color channels in the conventional method may not hold in the complicated environment with large head movements. However, Due to the robustness of the conventional methods under varying lighting conditions, conventional methods are more suitable in scenarios with less head movement, such as liveness detection in the payment systems.

\subsection{Improvement exploration to illumination variation}
We examined how brightness augmentation and image enhancement techniques improve the model's performance in various lighting conditions.
According to the results presented above, we can see that using image enhancement techniques cannot directly improve the HR estimation accuracy of deep learning-based methods. 
Despite these techniques eﬀectively improving image contrast and brightness, the accuracy gains achieved by tested models are quite limited. Some image enhancement methods even degrade estimated accuracy by disrupting frame temporal consistency and pixel value distribution. 
Further research is needed on image enhancement techniques for rPPG tasks.
From the results shown in Table \ref{brightness}, integrating with brightness augmentation leads to promising performance improvement on BH-rPPG, which is a powerful tool to force deep learning-based model to learn the feature invariance to lighting intensity.

In general, HR estimation tasks using rPPG are similar to other video-based learning tasks, such as action recognition, where the context of human anatomy is modeled between frames. The distinction is that rPPG modeled the color variation associated with HR. However, poor weather conditions introduce photometric differences that lead to degraded performance of the action recognition system. This is similar to the poor performance of the deep learning-based rPPG approach under different light intensities. Data augmentation is a straightforward and effective technique to remedy this deficiency in the field of action recognition. Inspired by this, we apply the data augmentation technique to deep learning-based HR estimation tasks. Experimental results demonstrate its effectiveness for rPPG tasks under different illumination conditions.

Compared to the image enhancement methods, the brightness augmentation improves the robustness of deep learning models under different lighting conditions. We argue that it is because the model without augmentation can be easily over-fitted to a specific range of pixel values. The image enhancement disrupted the original color distribution along both spatial and temporal dimensions, which is critical to rPPG signals. The brightness augmentation increased the data diversity and encouraged the model to focus on the features related to HR, which is invariant under different lighting conditions. 
However, most existing work of data augmentation focuses on single images rather than videos. Designing more useful data augmentation techniques can bridge the illumination robustness gap between deep learning-based methods and traditional methods. 

\section{Conclusion}
\label{sec:conclusion}
In this work, we compared the performance of different methods for rPPG-based heart rate estimation under three lighting intensities. The results show that conventional methods are more robust to lighting intensities changes and uneven lighting distribution, while the Physnet achieves the best performance among the deep learning-based methods. In the development of deep learning-based method, we should consider the varying lighting conditions, especially the different lighting intensities and uneven lighting distribution. Moreover, we conduct a comparative evaluation of deep learning-based techniques under the same training paradigms. The results show that Physnet achieves the best within UBFC-rPPG dataset and deep learning-based methods are able to capture the temporal variation of skin color with motion. 
Furthermore, we explore the potential method for performance enhancement to different illuminations. Results show that brightness augmentation is effective in improving performance in different lighting conditions.
The findings of this study urge additional research into developing more robust deep learning models to enable the actual application in daily living.

\ifCLASSOPTIONcaptionsoff
  \newpage
\fi

\bibliographystyle{IEEEtran}
\bibliography{rPPG}

\begin{thebibliography}{10}
\providecommand{\url}[1]{#1}
\csname url@samestyle\endcsname
\providecommand{\newblock}{\relax}
\providecommand{\bibinfo}[2]{#2}
\providecommand{\BIBentrySTDinterwordspacing}{\spaceskip=0pt\relax}
\providecommand{\BIBentryALTinterwordstretchfactor}{4}
\providecommand{\BIBentryALTinterwordspacing}{\spaceskip=\fontdimen2\font plus
\BIBentryALTinterwordstretchfactor\fontdimen3\font minus
  \fontdimen4\font\relax}
\providecommand{\BIBforeignlanguage}[2]{{%
\expandafter\ifx\csname l@#1\endcsname\relax
\typeout{** WARNING: IEEEtran.bst: No hyphenation pattern has been}%
\typeout{** loaded for the language `#1'. Using the pattern for}%
\typeout{** the default language instead.}%
\else
\language=\csname l@#1\endcsname
\fi
#2}}
\providecommand{\BIBdecl}{\relax}
\BIBdecl

\bibitem{statemonitoring}
Y.~Benezeth, P.~Li, R.~Macwan, K.~Nakamura, R.~Gomez, and F.~Yang, ``Remote
  heart rate variability for emotional state monitoring,'' in \emph{2018 IEEE
  EMBS International Conference on Biomedical \& Health Informatics
  (BHI)}.\hskip 1em plus 0.5em minus 0.4em\relax IEEE, 2018, pp. 153--156.

\bibitem{driver}
Y.-C. Tsai, P.-W. Lai, P.-W. Huang, T.-M. Lin, and B.-F. Wu, ``Vision-based
  instant measurement system for driver fatigue monitoring,'' \emph{IEEE
  Access}, vol.~8, pp. 67\,342--67\,353, 2020.

\bibitem{FaceLivenessDetection}
\BIBentryALTinterwordspacing
Face {{Liveness Detection}} by {{rPPG Features}} and {{Contextual
  Patch}}-{{Based CNN}} | {{Proceedings}} of the 2019 3rd {{International
  Conference}} on {{Biometric Engineering}} and {{Applications}}. [Online].
  Available: \url{https://dl.acm.org/doi/abs/10.1145/3345336.3345345}
\BIBentrySTDinterwordspacing

\bibitem{6523142}
G.~{de Haan} and V.~{Jeanne}, ``Robust pulse rate from chrominance-based
  rppg,'' \emph{IEEE Transactions on Biomedical Engineering}, vol.~60, no.~10,
  pp. 2878--2886, 2013.

\bibitem{chenDeepphysVideobasedPhysiological2018}
W.~Chen and D.~McDuff, ``Deepphys: Video-based physiological measurement using
  convolutional attention networks,'' in \emph{Proceedings of the European
  Conference on Computer Vision (ECCV)}, 2018, pp. 349--365.

\bibitem{iphys}
D.~McDuff and E.~Blackford, ``iphys: An open non-contact imaging-based
  physiological measurement toolbox,'' in \emph{2019 41st Annual International
  Conference of the IEEE Engineering in Medicine and Biology Society
  (EMBC)}.\hskip 1em plus 0.5em minus 0.4em\relax IEEE, 2019, pp. 6521--6524.

\bibitem{spetlikVisualHeartRate2018}
R.~{\v{S}}petl{\'\i}k, V.~Franc, and J.~Matas, ``Visual heart rate estimation
  with convolutional neural network,'' in \emph{Proceedings of the British
  Machine Vision Conference, Newcastle, UK}, 2018, pp. 3--6.

\bibitem{liRemoteHeartRate2014}
X.~Li, J.~Chen, G.~Zhao, and M.~Pietikainen, ``Remote heart rate measurement
  from face videos under realistic situations,'' in \emph{Proceedings of the
  IEEE conference on computer vision and pattern recognition}, 2014, pp.
  4264--4271.

\bibitem{yuRemoteHeartRate2019}
Z.~Yu, W.~Peng, X.~Li, X.~Hong, and G.~Zhao, ``Remote heart rate measurement
  from highly compressed facial videos: an end-to-end deep learning solution
  with video enhancement,'' in \emph{Proceedings of the IEEE/CVF International
  Conference on Computer Vision}, 2019, pp. 151--160.

\bibitem{verkruysseRemotePlethysmographicImaging2008}
W.~Verkruysse, L.~O. Svaasand, and J.~S. Nelson, ``Remote plethysmographic
  imaging using ambient light.'' \emph{Optics express}, vol.~16, no.~26, pp.
  21\,434--21\,445, 2008.

\bibitem{NoncontactAutomatedCardiac}
M.-Z. Poh, D.~J. McDuff, and R.~W. Picard, ``Advancements in noncontact,
  multiparameter physiological measurements using a webcam,'' \emph{IEEE
  transactions on biomedical engineering}, vol.~58, no.~1, pp. 7--11, 2010.

\bibitem{wangAlgorithmicPrinciplesRemote2016}
W.~Wang, A.~C. den Brinker, S.~Stuijk, and G.~De~Haan, ``Algorithmic principles
  of remote ppg,'' \emph{IEEE Transactions on Biomedical Engineering}, vol.~64,
  no.~7, pp. 1479--1491, 2016.

\bibitem{wangRobustAutomaticRemote2017}
W.~Wang, ``Robust and automatic remote photoplethysmography,'' 2017.

\bibitem{papageorgiou2014adaptive}
A.~Papageorgiou and G.~de~Haan, ``Adaptive gain tuning for robust remote pulse
  rate monitoring under changing light conditions,'' \emph{Technische
  Universiteit Eindhoven}, 2014.

\bibitem{PCA}
M.~Lewandowska, J.~Rumi{\'n}ski, T.~Kocejko, and J.~Nowak, ``Measuring pulse
  rate with a webcam—a non-contact method for evaluating cardiac activity,''
  in \emph{2011 federated conference on computer science and information
  systems (FedCSIS)}.\hskip 1em plus 0.5em minus 0.4em\relax IEEE, 2011, pp.
  405--410.

\bibitem{yuRemotePhotoplethysmographSignal2019}
Z.~Yu, X.~Li, and G.~Zhao, ``Remote photoplethysmograph signal measurement from
  facial videos using spatio-temporal networks,'' \emph{arXiv preprint
  arXiv:1905.02419}, 2019.

\bibitem{zhan2020analysis}
Q.~Zhan, W.~Wang, and G.~de~Haan, ``Analysis of cnn-based remote-ppg to
  understand limitations and sensitivities,'' \emph{Biomedical optics express},
  vol.~11, no.~3, pp. 1268--1283, 2020.

\bibitem{liOBFDatabaseLarge2018}
X.~Li, I.~Alikhani, J.~Shi, T.~Seppanen, J.~Junttila, K.~Majamaa-Voltti,
  M.~Tulppo, and G.~Zhao, ``The obf database: A large face video database for
  remote physiological signal measurement and atrial fibrillation detection,''
  in \emph{2018 13th IEEE International Conference on Automatic Face \& Gesture
  Recognition (FG 2018)}.\hskip 1em plus 0.5em minus 0.4em\relax IEEE, 2018,
  pp. 242--249.

\bibitem{balakrishnanDetectingPulseHead2013}
G.~Balakrishnan, F.~Durand, and J.~Guttag, ``Detecting pulse from head motions
  in video,'' in \emph{Proceedings of the IEEE Conference on Computer Vision
  and Pattern Recognition}, 2013, pp. 3430--3437.

\bibitem{ImprovedPulseDetection}
\BIBentryALTinterwordspacing
Improved pulse detection from head motions using {{DCT}} | {{IEEE Conference
  Publication}} | {{IEEE Xplore}}. [Online]. Available:
  \url{https://ieeexplore.ieee.org/abstract/document/7295070}
\BIBentrySTDinterwordspacing

\bibitem{takanoHeartRateMeasurement2007}
C.~Takano and Y.~Ohta, ``Heart rate measurement based on a time-lapse image,''
  \emph{Medical engineering \& physics}, vol.~29, no.~8, pp. 853--857, 2007.

\bibitem{allenPhotoplethysmographyItsApplication2007}
J.~Allen, ``Photoplethysmography and its application in clinical physiological
  measurement,'' \emph{Physiological measurement}, vol.~28, no.~3, p.~R1, 2007.

\bibitem{lamRobustHeartRate2015}
A.~Lam and Y.~Kuno, ``Robust heart rate measurement from video using select
  random patches,'' in \emph{Proceedings of the IEEE International Conference
  on Computer Vision}, 2015, pp. 3640--3648.

\bibitem{niuRhythmnetEndtoendHeart2019}
X.~Niu, S.~Shan, H.~Han, and X.~Chen, ``Rhythmnet: End-to-end heart rate
  estimation from face via spatial-temporal representation,'' \emph{IEEE
  Transactions on Image Processing}, vol.~29, pp. 2409--2423, 2019.

\bibitem{qiuEVMCNNRealtimeContactless2018}
Y.~Qiu, Y.~Liu, J.~Arteaga-Falconi, H.~Dong, and A.~El~Saddik, ``Evm-cnn:
  Real-time contactless heart rate estimation from facial video,'' \emph{IEEE
  transactions on multimedia}, vol.~21, no.~7, pp. 1778--1787, 2018.

\bibitem{wuEulerianVideoMagnification2012}
H.-Y. Wu, M.~Rubinstein, E.~Shih, J.~Guttag, F.~Durand, and W.~Freeman,
  ``Eulerian video magnification for revealing subtle changes in the world,''
  \emph{ACM transactions on graphics (TOG)}, vol.~31, no.~4, pp. 1--8, 2012.

\bibitem{MetarPPGRemoteHeart}
\BIBentryALTinterwordspacing
Meta-{{rPPG}}: {{Remote Heart Rate Estimation Using}} a {{Transductive
  Meta}}-learner | {{SpringerLink}}. [Online]. Available:
  \url{https://link.springer.com/chapter/10.1007/978-3-030-58583-9_24}
\BIBentrySTDinterwordspacing

\bibitem{VideoBasedRemotePhysiological}
\BIBentryALTinterwordspacing
Video-{{Based Remote Physiological Measurement}} via {{Cross}}-{{Verified
  Feature Disentangling}} | {{SpringerLink}}. [Online]. Available:
  \url{https://link.springer.com/chapter/10.1007/978-3-030-58536-5_18}
\BIBentrySTDinterwordspacing

\bibitem{sun2021camera}
Y.~Sun, J.~Hu, W.~Wang, M.~He, and P.~H. de~With, ``Camera-based discomfort
  detection using multi-channel attention 3d-cnn for hospitalized infants,''
  \emph{QUANTITATIVE IMAGING IN MEDICINE AND SURGERY}, vol.~11, no.~7, pp.
  3059--3069, 2021.

\bibitem{linusing}
X.~Lin and G.~de~Haan, ``Using blood volume pulse vector to extract rppg signal
  in infrared spectrum.''

\bibitem{smithTelehealthGlobalEmergencies2020}
A.~C. Smith, E.~Thomas, C.~L. Snoswell, H.~Haydon, A.~Mehrotra, J.~Clemensen,
  and L.~J. Caffery, ``Telehealth for global emergencies: Implications for
  coronavirus disease 2019 (covid-19),'' \emph{Journal of telemedicine and
  telecare}, vol.~26, no.~5, pp. 309--313, 2020.

\bibitem{huang2020heart}
P.-W. Huang, B.-J. Wu, and B.-F. Wu, ``A heart rate monitoring framework for
  real-world drivers using remote photoplethysmography,'' \emph{IEEE journal of
  biomedical and health informatics}, vol.~25, no.~5, pp. 1397--1408, 2020.

\bibitem{fernandesPredictingHeartRate2019}
S.~Fernandes, S.~Raj, E.~Ortiz, I.~Vintila, M.~Salter, G.~Urosevic, and S.~Jha,
  ``Predicting heart rate variations of deepfake videos using neural ode,'' in
  \emph{Proceedings of the IEEE/CVF International Conference on Computer Vision
  Workshops}, 2019, pp. 0--0.

\bibitem{ubfc-rppg}
\BIBentryALTinterwordspacing
Unsupervised skin tissue segmentation for remote photoplethysmography -
  {{ScienceDirect}}. [Online]. Available:
  \url{https://www.sciencedirect.com/science/article/pii/S0167865517303860}
\BIBentrySTDinterwordspacing

\bibitem{VIPL}
X.~Niu, H.~Han, S.~Shan, and X.~Chen, ``Vipl-hr: A multi-modal database for
  pulse estimation from less-constrained face video,'' in \emph{Asian
  Conference on Computer Vision}.\hskip 1em plus 0.5em minus 0.4em\relax
  Springer, 2018, pp. 562--576.

\bibitem{PURE}
R.~Stricker, S.~M{\"u}ller, and H.-M. Gross, ``Non-contact video-based pulse
  rate measurement on a mobile service robot,'' in \emph{The 23rd IEEE
  International Symposium on Robot and Human Interactive Communication}.\hskip
  1em plus 0.5em minus 0.4em\relax IEEE, 2014, pp. 1056--1062.

\bibitem{MAHNOB}
M.~Soleymani, J.~Lichtenauer, T.~Pun, and M.~Pantic, ``A multimodal database
  for affect recognition and implicit tagging,'' \emph{IEEE transactions on
  affective computing}, vol.~3, no.~1, pp. 42--55, 2011.

\bibitem{cohface}
\BIBentryALTinterwordspacing
G.~Heusch, A.~Anjos, and S.~Marcel. A {{Reproducible Study}} on {{Remote Heart
  Rate Measurement}}. [Online]. Available:
  \url{http://arxiv.org/abs/1709.00962}
\BIBentrySTDinterwordspacing

\bibitem{KLT}
C.~Tomasi and T.~Kanade, ``Detection and tracking of point,'' features.
  Technical Report CMU-CS-91-132, Carnegie, Mellon University, Tech. Rep.,
  1991.

\bibitem{guo2020zero}
C.~Guo, C.~Li, J.~Guo, C.~C. Loy, J.~Hou, S.~Kwong, and R.~Cong,
  ``Zero-reference deep curve estimation for low-light image enhancement,'' in
  \emph{Proceedings of the IEEE/CVF Conference on Computer Vision and Pattern
  Recognition}, 2020, pp. 1780--1789.

\bibitem{paszke2019pytorch}
A.~Paszke, S.~Gross, F.~Massa, A.~Lerer, J.~Bradbury, G.~Chanan, T.~Killeen,
  Z.~Lin, N.~Gimelshein, L.~Antiga \emph{et~al.}, ``Pytorch: An imperative
  style, high-performance deep learning library,'' \emph{Advances in neural
  information processing systems}, vol.~32, pp. 8026--8037, 2019.

\bibitem{tulyakovSelfadaptiveMatrixCompletion2016}
S.~Tulyakov, X.~Alameda-Pineda, E.~Ricci, L.~Yin, J.~F. Cohn, and N.~Sebe,
  ``Self-adaptive matrix completion for heart rate estimation from face videos
  under realistic conditions,'' in \emph{Proceedings of the IEEE conference on
  computer vision and pattern recognition}, 2016, pp. 2396--2404.

\bibitem{lee2020meta}
E.~Lee, E.~Chen, and C.-Y. Lee, ``Meta-rppg: Remote heart rate estimation using
  a transductive meta-learner,'' in \emph{European Conference on Computer
  Vision}.\hskip 1em plus 0.5em minus 0.4em\relax Springer, 2020, pp. 392--409.

\bibitem{taylor2014adaptive}
M.~J. Taylor and T.~Morris, ``Adaptive skin segmentation via feature-based face
  detection,'' in \emph{Real-Time Image and Video Processing 2014}, vol.
  9139.\hskip 1em plus 0.5em minus 0.4em\relax International Society for Optics
  and Photonics, 2014, p. 91390P.

\bibitem{wangExploitingSpatialRedundancy2014}
W.~Wang, S.~Stuijk, and G.~De~Haan, ``Exploiting spatial redundancy of image
  sensor for motion robust {{rPPG}},'' vol.~62, no.~2, pp. 415--425.

\end{thebibliography}

\end{document}